# Incrementality Bidding & Attribution[1]

By Randall Lewis, Netflix
Jeffrey Wong, Netflix

First Draft: 2017/04/21
This Draft: 2018/03/08

[1] The author(s) would like to thank the valuable contributions of many at Netflix to this research: management & ad operations (Gagan Hasteer, Kelly Uphoff, Steve McBride, Michael Pow, James Ouska), ad tech engineering (Duo Wang, Kai Hu, Raghu Srinivasan, Stephen Walz), and science & algorithms (Vijay Bharadwaj, Benoit Rostykus, Tony Jebara, Dave Hubbard).

# Abstract

The causal effect of showing an ad to a potential customer versus not, commonly referred to as "incrementality," is the fundamental question of advertising effectiveness. In digital advertising three major puzzle pieces are central to rigorously quantifying advertising incrementality: ad buying/bidding/pricing, attribution, and experimentation. Building on the foundations of machine learning and causal econometrics, we propose a methodology that unifies these three concepts into a computationally viable model of both bidding and attribution which spans the randomization, training, cross validation, scoring, and conversion attribution of advertising's causal effects. Implementation of this approach is likely to secure a significant improvement in the return on investment of advertising.

# Introduction

Measuring the causal effects of advertising has traditionally been difficult. Standard methods to evaluate digital advertising effectiveness rely on correlational and proxy metrics such as click- or view-through conversions, which work quite well for some advertisers, but much more poorly for others, especially large brands.

To evaluate the standard industry methods, researchers have employed the scientific method in the past by running experiments, subject to economic, statistical, and technological constraints discussed by Lewis & Rao (2015). In spite of these constraints, researchers at eBay discovered a massive misallocation of that company's resources due to the use of correlational metrics such as click-through conversion attribution guiding advertising budget allocation decisions.

Other researchers at Yahoo and Google identified cases where advertising had statistically significant and even potentially profitable effects on sales (Johnson, Lewis, & Reiley 2016; Johnson, Lewis, & Nubbemeyer 2017).

That said, these experiments represent outliers to "business as usual," due to the financial and technical costs of designing, configuring, executing, analyzing, and evaluating advertising experiments. Most advertisers do not employ scientists to run large-scale field experiments to evaluate their advertising's return on investment (ROI)---they use the standard tools and services offered by ad technology companies. These companies operate automated ad-buying technologies that help advertisers optimize their advertising spend in line with the metrics that the ad technology companies can track on behalf of the advertisers. However, the optimization methods are inherently correlational due to the statistical methodology or metrics utilized. In short, these tools "optimize for the 'wrong' metric."

Attribution models have emerged in an attempt to correct or realign the "wrong" metrics. By applying "last-touch" or "last-view" attribution models, advertisers who are skeptical of the correlation between clicks and conversions can help remove the distortions imposed by ad-buying technologies which optimize for just click-through conversions. However, in the end, advertisers do not care about ad clicks or impressions---they want to know, "Does my advertising work?" This is a question about the causal or incremental effect of advertising and, hence, requires causal modeling.

We propose a method that encompasses the ad impression valuation/bidding/buying and attribution problem using rigorous causal methods powered by randomization mechanisms, similar to Johnson, Lewis, & Nubbemeyer's (2017) "Predicted Ghost Ads." The method can be simply summarized: we use a regularized instrumental variable (IV) model of heterogeneous treatment effects in a continuous-time panel. While most of the components of the model may not be novel, the viability of their combination represents a step forward for incrementality bidding.

# Introduction to Incrementality

In this section we provide a basic introduction to incrementality by using the simplest model possible to define all of the core concepts. We build upon this basic introduction in Advanced Incrementality for Industry where we present a generalization of this model that addresses practical issues that arise in real-world dynamic real-time bidding and attribution models.

## Defining Incrementality Attribution and Bidding

We begin with the simplest model possible:
$$y_i = \alpha + \beta x_i + \varepsilon_i.$$

Here, $i$ denotes a given user, $y_i$ is the outcome of interest (e.g., hereafter referred to as a "conversion" such as a visit to a website, transactions, or sales), $x_i$ is a count of ad exposures, and $\varepsilon_i$ is the random error term. The parameter $\alpha$ represents the conversion baseline in the absence of advertising; if $\alpha > 0$, users convert without seeing ads, underscoring the need to measure incrementality. The parameter $\beta$ represents the *incrementality*, or causal effect, attributable to each ad exposure in units of the conversion:

$$\Delta y_i = E[y_i|show\ ad] - E[y_i|don't\ show\ ad] = \beta.$$

If the incrementality is sufficiently large, advertising is a profitable investment. Naturally, breakeven requires a translation of conversions into currency which depends on the conversion's value $v$, gross profit margins $m$, and the cost of the advertising $c$: advertise if $\beta \cdot m \cdot v > c$. This profitability condition represents a simple ad valuation or bidding rule---never pay more for an ad than its contribution to gross profits. This can be rewritten in terms of the Cost Per Incremental Action (CPIA) pricing model described by Johnson & Lewis (2015):

$$CPIA = \frac{c}{\beta} < m \cdot v = \pi = Gross\ Profits\ per\ Conversion.$$

Its corresponding attribution rule is to give the *incrementality share*, $s_i = \frac{x_i\beta}{\alpha + x_i\beta}$, of credit for the conversion to be divided among all exposures. This overly simple model implicitly attributes an equal share of the total incrementality to each impression regardless of its characteristics or timing relative to the conversion---we will eventually remedy these flaws below.

Other commonly reported metrics include the *lift* over baseline caused by an impression defined as $\% = \frac{\beta}{\alpha}$ and the *conversion lift* of all ads over baseline defined as $\%_c = \frac{x_i\beta}{\alpha}$. These metrics are generally less useful for bidding and other business decisions because they are unit-free in relative to the economically irrelevant baseline $\alpha$. Specifically, recall that $\alpha$ does not appear in the profitability condition or bidding rule. Intuitively, if our ad campaign generates 1,000 incremental conversions, we should be indifferent to whether they represent a 100% or 10% lift, corresponding to baselines of 1,000 and 10,000 conversions, respectively, when evaluating the campaign. Hence, we recommend reporting metrics such as incremental conversions and CPIA for actionable decision-making.

| Name | Formula | Definition |
|---|---|---|
| Conversion Value | $v_i$ | Value (zero or otherwise) of user *i*'s conversion event. |
| Conversion Margin | $m_i$ | Gross profit margin of user *i*'s conversion event. |
| Conversion Profits | $\pi_i = m_i \cdot v_i$ | Gross profit of user *i*'s conversion event. |
| Cost per Exposure | $c$ | Cost of each impression shown to user *i*. |
| Number of Exposures | $x_i$ | Number of ads shown to user *i*. |
| Baseline Conversions | $\alpha$ | Conversion rate in the absence of advertising. |
| Ad Effect | $\beta$ | Incremental impact of an ad exposure on conversions. |
| Incremental Conversions | $\Delta y_i = x_i\beta$ | Total expected effect of ads on conversions by user *i*. |

| Lift per Impression | $\% = \frac{\beta}{\alpha}$ | Ad effect relative to baseline conversions. |
|---|---|---|
| Conversion Lift | $\%_c = \frac{\Delta y_i}{\alpha}$ | Total expected effect of ads relative to baseline conversions. |
| Incrementality Share | $s_i = \frac{\Delta y_i}{\alpha + \Delta y_i}$ | Expected share of user $i$'s conversions caused by the ads. |
| Incremental Profit | $\pi_i \cdot s_i$ | Incremental gross profits caused by ads shown to user $i$. |
| Cost per Action | $CPA = \frac{x_i c}{\alpha + \Delta y_i}$ | Total cost of user $i$'s ad exposures divided by the expected number of conversions. |
| Cost per Incremental Action | $CPIA = \frac{c}{\beta}$ | Cost of user $i$'s ad exposure divided by its ad effect. |

## Optimizing Incrementality through Attribution and Bidding

The simple model above answers the question, "How much should I be willing to pay for advertising?" However, advertising is not homogeneous, and hence, the answer to optimizing ad-buying for incrementality is more complex. To accommodate heterogeneous advertising, we introduce $W$ as a characteristic (i.e., context) of the advertising:

$$y_i = \alpha(W) + \sum_k \beta_k x_{ik} + \varepsilon_i.$$

$$x_{ik} = \sum_{j \in impressions} x_{ijk}; \ x_{ijk} \equiv 1(impression \ j \ has \ characteristic \ k)$$

Here, $W$ is either a characteristic of the user (e.g., country or state) or of the ad (e.g., size, placement, or creative). The parameter $\alpha(W)$ represents the baseline as it varies with the distribution of characteristics. The next term of the equation presents the total incrementality over various types of advertising, each with its own magnitude of effect on conversions. The total incrementality breaks down into the contributions by each ad to each characteristic. The characteristic effect, $\beta_k$, represents the amount of incrementality attributable to an ad possessing characteristic $k$; impression $j$'s ad effect is the sum over its characteristics' attributed incrementality:

$$\Delta y_{ij} = \sum_k \beta_k x_{ijk}.$$

For each impression, we can again construct a simple valuation-based bidding rule, $b_{ij} = mv \cdot \sum_k \beta_k x_{ijk}$, with its corresponding ROI: $r_w \equiv \frac{b_{ij}}{c_{ij}} - 1 > 0$, given an impression cost $c_{ij}$. ROI can be compared across all types of advertising investments, enabling a constrained marketing budget to prioritize those with the highest ROI. For example, suppose we paid $5 cost per mille (CPM, cost per thousand impressions) for impressions in Canada and the USA; however, we learned that Canada yielded $10 of gross profit whereas the USA yielded only $5 per mille. We could increase total profits by allocating a larger share of advertising expenditures to Canada and a correspondingly smaller share to the USA.

Incrementality share and lift metrics have straightforward generalizations, shown below.

| Name | Formula | Definition |
|------|---------|------------|
| Cost per Exposure | $c_{ij}$ | Cost of impression $j$ shown to user $i$. |
| Number of Exposures | $x_{ik}$ | Number of ads with characteristic $k$ shown to user $i$. |
| Baseline Conversions | $\alpha(W)$ | The conversion rate in the absence of advertising. |
| Characteristic Effect | $\beta_k$ | The attribution of characteristic $k$ to the incremental impact of an ad exposure. |
| Incremental Conversions | $\Delta y_i = \sum_k \beta_k x_{ik}$ | The total effect of ads on user $i$. |
| Ad Effect | $\Delta y_{ij} = \sum_k \beta_k x_{ijk}$ | The effect of impression $j$ shown to user $i$. |
| Lift per Impression | $\%_j = \frac{\Delta y_{ij}}{\alpha(W)}$ | The ad effect relative to baseline conversions. |
| Conversion Lift | $\%_c = \frac{\Delta y_i}{\alpha(W)}$ | The total effect of ads relative to baseline conversions. |
| Incrementality Share | $s_i = \frac{\Delta y_i}{\alpha(W) + \Delta y_i}$ | The share of user $i$'s conversions caused by the ads. |
| Cost per Incremental Action | $CPIA = \frac{c_{ij}}{\Delta y_{ij}}$ | Cost of user $i$'s ad exposure divided by its ad effect. |

# Estimating Incrementality

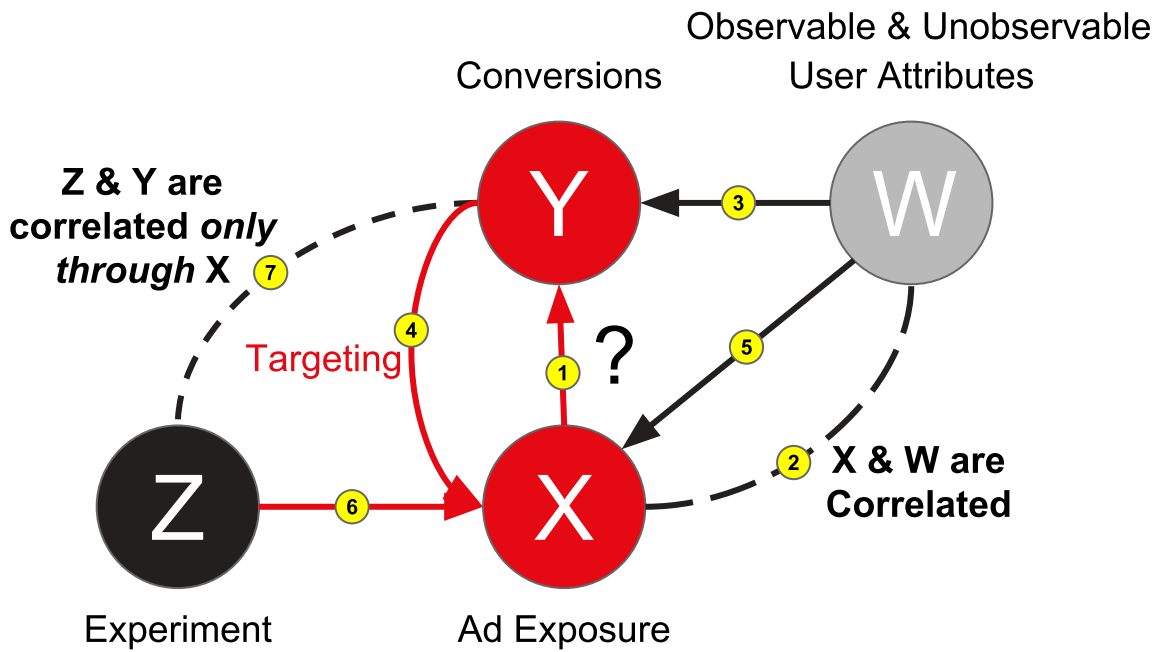

Figure 1 - Causal graph of ad effectiveness measurement

We have looked at two simple regression models which incorporate incrementality, but the real world of advertising is more complex, as diagrammed in Figure 1. While the models make it look simple to estimate incrementality, the causal effect of advertising (Fig. 1, arrow 1), with an ordinary least squares (OLS) regression (or any similar tabulation method), advertising exposure does not satisfy the assumptions required for OLS to be unbiased---specifically the assumption of conditional exogeneity of the ad exposure regressors, $x_{iw}$. This is the fundamental problem of measuring the causal effect of targeted advertising: ad exposure is not randomized (Fig. 1, arrow 2). In digital advertising, this arises for several reasons: prospects differ in their conversion rate (Fig. 1, arrow 3), marketers explicitly target their best prospects (Fig. 1, arrows 4 & 5), advertisers compete for common prospects (Fig. 1, arrow 5), and prospects choose when to make themselves eligible for advertising exposure (e.g., "activity bias") (Fig. 1, arrow 5). These are all forms of selection bias that lead naive estimates of incrementality to be biased---and in practice, severely so (e.g., Lewis, Rao, & Reiley 2011; Blake, Nosko, & Tadelis 2014). Selection bias is one source of endogeneity, a well-understood challenge in causal econometrics with known solutions. Randomized experimentation that influences ad exposure (Fig. 1, arrow 6) solves this problem by constructing estimates that are decoupled from all of the sources of bias (Fig. 1, arrow 7). We formalize this diagram using the standard causal method of instrumental variables (IV).

We return to our first model to illustrate:
$$y_i = \alpha + \beta x_i + \varepsilon_i.$$
Given the potential severity of selection bias in digital advertising settings (Lewis & Rao 2015), we advocate the use of instrumental variables (IVs) obtained through randomized experiments. In the section on <u>User- versus Bid-Level Randomization: Tradeoffs</u>, we provide more in-depth discussion as to why bid-level randomization provides better causal guarantees and, accordingly, systems that are easier to build, manage, and maintain. Intuitively, randomizing by bid ensures that we can causally disentangle the difference between the types of ads seen by users who are responsive to ads and the particular types of ads that are best at inducing responses for each type of user. However, here we discuss a simple example using user-level randomization. The example presents a simple multiple treatment generalization of the local average treatment effect (LATE) (Angrist & Imbens 1994).

Suppose we randomly assign a user to either the treatment group ($Z = 1$) or control ($Z = 0$). We can then construct a user-level instrumental variable for exposure, $z_i \equiv Z \cdot a_i$, where $a_i$ represents the number of user $i$'s auctions. The instrument $z_i$ conditionally satisfies the exogeneity assumption, $E[z_i'\varepsilon_i | a_i] = 0$, required for IV estimation to be consistent as long as we account for the context $W$ of the (potentially unobservable counterfactual) number of ad auctions, $a_i$, for which user $i$ is eligible if assigned to the treatment group (Lewis 2014; Johnson, Lewis, and Reiley 2015).

To summarize our new IV model:
$$y_i = \alpha_0 + \alpha_1 a_i + \beta x_i + \varepsilon_i \text{ ("Second Stage")}$$
$$x_i = \pi_0 + \pi_1 a_i + \pi_z z_i + \nu_i \text{ ("First Stage")}$$

Here, the $\pi$ parameters model the number of exposures as a function of the randomization $z_i$ and other factors such as the number of auctions, $a_i$. In particular, $\pi_z$ represents the auction "win rate" of users in the treatment group rather than control. In this model, we define the baseline with the simple functional form of $\alpha(W) = \alpha_0 + \alpha_1 a_i$. Johnson, Lewis, & Reiley (2015) discuss several variants of this model both with and without instrumental variables.

This model forms the foundation of our causal estimation and optimization by accounting for the number of auctions for each type of impression characteristic:

$$y_i = \alpha(W) + \sum_k \beta_k x_{ik} + \varepsilon_i \quad \text{(“Second Stage”)}$$

$$x_{ik} = \pi_{0,k}(W) + \sum_{k'} \pi_{k'} z_{ik'} + \nu_{ik} \quad \text{(“First Stage”)}$$

Here, $W$ includes the number of auctions for all types of ad characteristics. We estimate both the baseline conversion rate, $\alpha(W)$, and the baseline win rate, $\pi_0(W)$. $x_{ik}$ and $z_{ik}$ reflect the number of user $i$'s impressions and auctions with characteristic $k$; $\pi_{k'}$ estimates heterogeneous win rates across auctions with characteristics $k$.

These IV models can be estimated using standard econometric techniques such as 2-stage least squares (2SLS) or its nonlinear generalization, the generalized method of moments (GMM). We center our discussion here and throughout the remainder of the paper on linear, additive 2SLS for both conceptual and computational simplicity.

# Advanced Incrementality for Industry

In the previous section, we introduced the reader to a simple exposition of the core concepts needed to understand incrementality: heterogeneous ad effects, incrementality share, and instrumental variables. In this section, we expound on a number of practical considerations to make incrementality a tractable causal machine learning decision problem.

## Background: Practical Requirements for a Real-Time Bidder

When developing a method to address incrementality bidding and attribution, we need to not only satisfy the theoretical ideal of causal estimation, but also balance the practical engineering constraints of participating in the real-time bidding (RTB) marketplace and efficiency requirements of extracting the most value, both statistical and economic, from each purchased impression.

The engineering constraints of operating a bidder and providing attribution reporting deal with data volume, computational load, and time constraints. For example, data processing needs to accommodate >10 billion auctions per day and construct training data for the bidding and attribution model to estimate or approximate the behavior of >1 billion users per month. Both bidding and attribution modeling need to be capable of valuing any possible combination of ad impression and conversion behavior; hence, such values need to be decomposable by the contributions from user, impression, and ad characteristics. The bid calculation must be very fast (e.g., less than 10 milliseconds) in order to ensure participation in the ad auctions. Model training must be fast enough to compute or update multiple times per hour. Finally, the model must account for sequential feedback induced by human or algorithmic decisions which resulted from prior bidding behavior---if a user responds to a prior ad or not, the model must be robust to that endogenous feedback within seconds, if not milliseconds.

Efficiency requires extracting the most value from each purchased impression. Statistically, we must maximize the signal-to-noise to reduce the variance within the aforementioned practical computational constraints. Simple methods that include observations that most reduce uncertainty can help. Additionally, performing attribution using post-impression information, such as viewability, can improve power, though bids can only use pre-impression information, such as expected viewability. Finally, the statistical models need reliable and timely feedback on both the model's predictions and realized incrementality and estimates of their respective uncertainty. Quantifying the uncertainty is especially important since ad effects tend to be small and statistically imprecise (Lewis & Rao 2015).

For economic efficiency, we must avoid biases in our bidding and attribution (e.g., over- or under-bidding) and cost-effectively introduce randomization (e.g., randomize bidding only to resolve uncertainty to reduce over- and under-bidding). A cost-effective bidding method should reflect our statistical uncertainty of the value of each ad by sampling from the quasi-posterior of our covariance matrix. Further, using causal methods only when they are required can help reduce excessive exploration---if evidence points toward a correlational model being consistent with the causal estimates, then barring any strong argument against the correlational model, using it can reduce the errors made from over-exploration, perhaps at the cost of some bias.

With this set of constraints and challenges in mind, we proceed to our dynamic incrementality bidding and attribution model.

## Ad Stock: Impression Features for Bidding & Scoring

An incrementality model evaluates each impression. Those valuations can be used online in real-time bidding (RTB) or for post-bid offline advertising attribution. We propose using a flexible ad stock model to account for natural variation in incrementality due to differences in the delay between ad exposure and conversion events.

### Modeling Ad Stock in Continuous Time

We return to our first model to introduce the real-world complexity of continuous time:

$$y_i(t) = \alpha(t) + \beta x_i(t) + \varepsilon_i(t).$$

This continuous-time model describes the conversion outcome in terms of expected rates.[2] For example, $\alpha(t)$ describes how the baseline sign-up rate evolves over time while $x_i(t)$ represents the level of ad stock for user $i$ at time $t$. Our first model, which abstracted away from time, can be represented as the time integral of this model.

We explicitly define *ad stock* in units of impressions. Ad stock is consumed over time in proportion to its effect on user behavior. For example, if an impression increases sales by 1% during the first 5 days after exposure, then each day consumes ⅕ of the impression's ad stock. Alternatively, if an impression increases sales by a total of $10 but has ad stock which experiences daily geometric decay of ½, then the first and second days consume ½ and ¼ of the impression's ad stock, respectively. In these two examples, respectively, $x_i(t) = \{1/5 \ for \ t \in [0,5], \ 0 \ elsewhere\}$ and $x_i(t) = \{2^{-ceiling(t)} \ for \ t > 0, \ 0 \ elsewhere\}$. These are conceptually simple examples of probability density functions, $f(t|\theta)$ where $\theta$ is a shape parameter, that can be used to model ad stock. While these examples are suitable for modeling ad stock in discrete time, they are less ideal for continuous-time modeling because they have discontinuities at date boundaries. Hence, for the remainder of the paper, we use continuous exponential and gamma distributions: $f(t|\tau) = \frac{1}{\tau}e^{-t/\tau}$ and $f(t|k,\theta) = \frac{1}{\Gamma(k)\theta^t}t^{k-1}e^{-t/\theta}$ where $\Gamma(k)$ is the canonical gamma function. In particular, the "memoryless" property of the exponential distribution is helpful for both analytic derivations and engineering complexity.

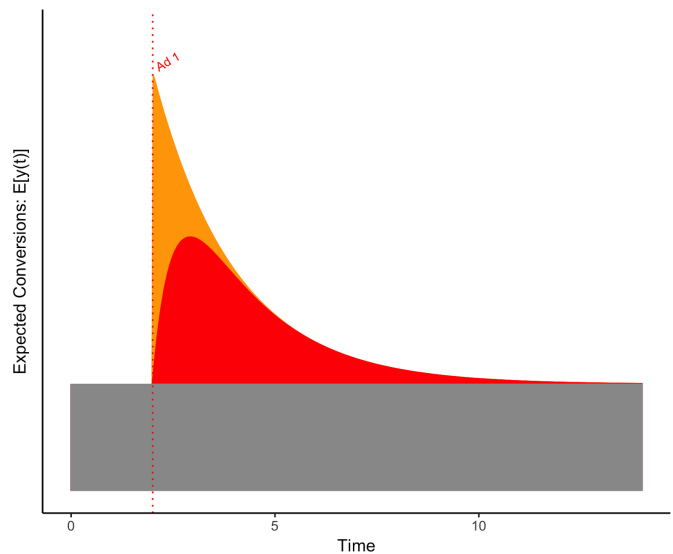

Figure 2 - Modeling ad stock as a mixture of two exponential distributions

A user's ad stock, $x_i(t)$, aggregates each impression's ad stock contribution:

---

$$x_i(t) = \sum_{j \in impressions} x_{ij}(t)$$

$$x_{ij}(t) = A_j \cdot f\left(t - t_j | \theta\right).$$

$A_j = 1(t > t_j)$ defines the one unit of ad stock contributed by impression $j$ at $t = t_j$. The ad stock then dissipates at the rate defined by $f\left(t - t_j | \theta\right)$. Although we do not know the true shape of $f\left(t - t_j | \theta\right)$ with respect to either $\theta$ or even $f(\cdot)$, we can model it as a mixture of distributions. We rewrite $x_{ij}(t)$ as $x_{ij}(t|\theta)$ and our mixture model as

$$y_i(t) = \alpha(t) + \sum_{\theta} \beta_{\theta} x_i(t|\theta) + \varepsilon_i(t).$$

To model heterogeneous effects of the ad impressions, richer features can be constructed by multiplying an impression's contribution to ad stock, $x_{ij}(t)$, by a predetermined weight, $w_{ij}$, to rescale the ad stock magnitude:

$$x_i(t) = \sum_{j \in impressions} w_{ij} \cdot x_{ij}(t).$$

For each unit of ad stock, $A_j$, these features now contribute $w_{ij}$ instead of 1. Weights can reflect the expected intensity of an ad effect using predetermined aggregations such as $w_{ij} = Pr(conversion \mid user\ visited\ impression\ j's\ domain)$, $w_{ij} = Pr(ad\ is\ viewable)$, or $w_{ij} = Pr(user\ i\ is\ a\ bot)$. More generally, weights operate as traditional impression or bid features, encoding information about the history of the user or ad characteristics such as number of ads shown in the past hour, country, ad size, domain, media type, or creative. To differentiate among these numerous weights, we append the index $k$ to $w_{ij}$ to obtain $w_{ijk}$. These weights represent the context $W_i(t)$ in terms of $K$ features at time $t$, which includes information about the current impression $j$ and can encode other features based on information available prior to $t$. Finally, $w_{ijk}$ can produce a hierarchy of features such as $x_{ij}(t)$ for which $w_{ijk} = 1$ and $x_{ijk}(t) \equiv w_{ijk} \cdot x_{ij}(t)$ for our heterogeneous treatment effects model in continuous time:

$$y_i(t) = \alpha(t|W) + \beta x_i(t) + \sum_{k} \beta_k x_{ik}(t) + \varepsilon_i(t).$$

Without loss of generality, we simplify notation by defining $k$ to index all weights $w_{ijk}$ that impression $j$ takes on, including the special case of constant weighting $w_{ij} = 1 \ \forall i, j$:

$$y_i(t) = \alpha(t|W) + \sum_{k} \beta_k x_{ik}(t) + \varepsilon_i(t).$$

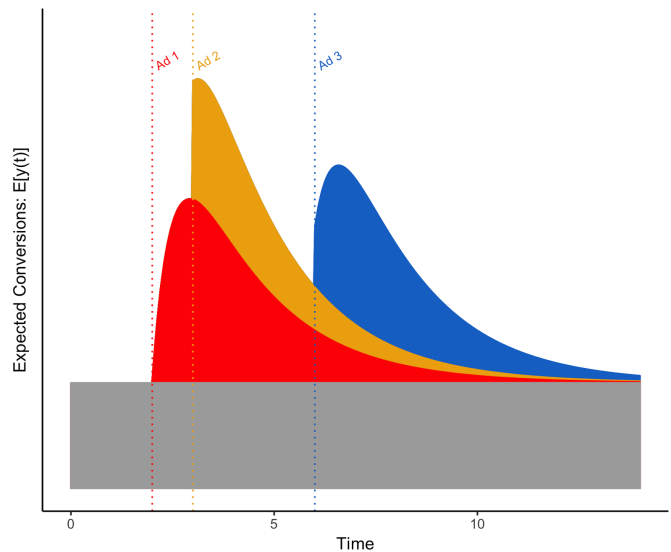

Figure 3 - Incrementality model with 3 heterogeneous ads

Combining our mixture and heterogeneous treatment effects models, we obtain our final model:

$$y_i(t) = \alpha(t|W) + \sum_k \sum_\theta \beta_{k,\theta} x_{ik}(t|\theta) + \varepsilon_i(t).$$

Our final formulation implies a tractable linear[3] model of incrementality in continuous time:

$$\Delta y_i(t) = \sum_k \sum_\theta \beta_{k,\theta} x_{ik}(t|\theta) = \sum_k \sum_\theta \beta_{k,\theta} \sum_j x_{ijk}(t|\theta) = \sum_j \sum_k \sum_\theta \beta_{k,\theta} w_{ijk} x_{ij}(t|\theta) = \sum_j \Delta y_{ij}(t).$$

In summary, we model user-level incrementality as the sum over each impression's contribution to each characteristic's effect, aided by an additive mixture model. This model readily calculates incrementality by user or impression and decomposes either by each impression characteristic---we can now do incrementality attribution.

---

[3] See <u>Appendix: Non-Additive Incremental Value</u> for a discussion of nonlinear models.

## Incrementality Attribution

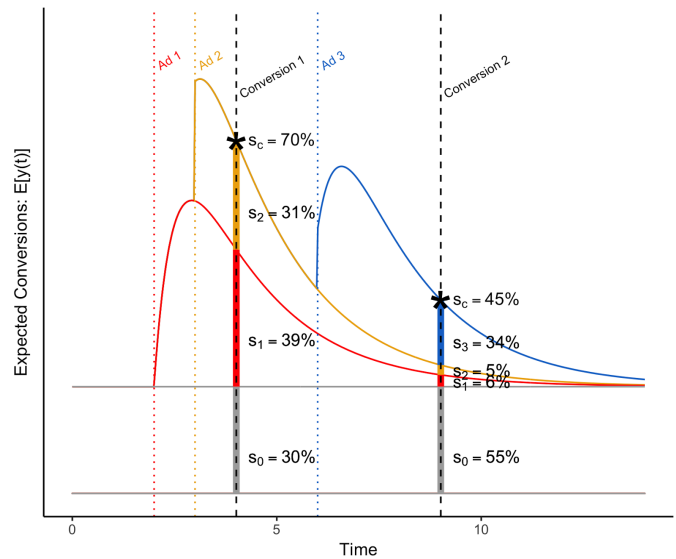

Figure 4 - Incrementality shares: 3 ads on 2 conversions

Our incrementality model assigns an incrementality share to each conversion and values each impression.

Each conversion's incrementality share can be computed as outlined above, except using the continuous-time model. We illustrate with a simple model:

$$s_{ic} = \frac{E[\Delta y_i(t_c)]}{E[y_i(t_c)]} = \frac{\beta x_i(t_c)}{\alpha(t_c) + \beta x_i(t_c)}$$

where $t_c$ denotes the time of the conversion $c$. In words, the incrementality share is the ratio of the expected ad effect to the expected outcome at the time of the conversion, including the ad effect.

Each impression $j$'s contribution to the incrementality of a given conversion is similar:

$$s_{ijc} = \frac{E[\Delta y_{ij}(t_c)]}{E[y_i(t_c)]} = \frac{\beta x_{ij}(t_c)}{\alpha(t_c) + \beta x_i(t_c)}.$$

The conversion's incrementality share equals the sum of the user's impressions' incrementality shares:

$$s_{ic} = \sum_{j \in impressions} s_{ijc}.$$

Upon aggregating over all of user $i$'s conversions that might have been influenced by impression $j$, $s_{ij} = \sum_{c \in conversions} s_{ijc}$, we can write:

$$s_i = \sum_{c \in conversions} s_{ic} = \sum_{j \in impressions} \sum_{c \in conversions} s_{ijc} = \sum_{j \in impressions} s_{ij}.$$

This leads to a convenient result: the total incremental effect of an ad campaign can be represented by scoring either its impressions or its conversions with incrementality shares and computing their sum.

$$Campaign\ Incrementality = \sum_{i \in users} s_i = \sum_{i \in users} \sum_{j \in impressions} s_{ij}.$$

The third way of computing the total incrementality of a campaign is to sum the expected incremental value (i.e., the ad effects) of all impressions in the campaign:

$$Campaign\ Incrementality = \sum_{i \in users} \sum_{j \in impressions} \Delta y_{ij}.$$

However, to estimate the expected incremental value of impression $j$, $\Delta y_{ij}$, we first need to discuss incrementality bidding.

## Incrementality Bidding

To buy ads cost effectively, we need a model to value their incremental effects before a conversion happens, in contrast to attributing value after the conversion occurs as with incrementality share. We need to estimate the unconditional difference in expectation which, thanks to the linear model, is quite simple (see [Appendix: Non-Additive Incremental Value](#) for nonlinear models):

$$\Delta y_{ij}(t) = E[y_i(t)|show\ ad\ j] - E[y_i(t)|don't\ show\ ad\ j] = \beta x_{ij}(t).$$

Because we cannot observe the timing of future conversions to compute an impression's incrementality share, we instead sum (i.e., integrate) the ad effect over time:

$$\Delta y_{ij} = \int_{t_j}^{t_j+\bar{T}} \Delta y_{ij}(t)dt = \int_{t_j}^{t_j+\bar{T}} \beta x_{ij}(t)dt = \beta \cdot A_j w_{ij} \int_0^{\bar{T}} f(t-t_j|\theta)dt = \beta \cdot w_{ij}.$$

The final equality shows that the incremental value of an impression is simply the coefficient $\beta$ times the impression's weight, $w_{ij}$. This follows from the use of a probability density function $f(t|\theta)$ which is defined to integrate to 1 on $[0, \bar{T}]$ and our definition of $A_j = 1(t > t_j)$ ---each impression contributes a single unit of ad stock. This trivial time integral implies a simple sum over coefficients and weights for the full incrementality model: $\Delta y_{ij} = \Sigma_k \Sigma_\theta \beta_{k,\theta} w_{ijk}$. Having a simple formula for the value of an ad is a requirement for real-time bidding in which bids must be computed and submitted within milliseconds. This model bypasses the burden of time integrals by pushing all time-related computations to the continuous-time ad stock variables.

We would use $\Delta y_{ij} = \Sigma_k \Sigma_\theta \beta_{k,\theta} w_{ijk}$ as an input into a strategic bidding model to bid for incrementality just as we would with any other ad-valuation model. For example, in a simple second-price auction, we could use the equilibrium strategy of "bidding our valuation" by submitting our $bid = \beta \cdot w_{ij}$. In a first-price auction, $\beta \cdot w_{ij}$ would be the upper bound on our willingness to pay for the ad; we would want to employ an optimal bid-shading rule based on our beliefs of the distribution of other advertisers' valuations of the ads. Hence, the bidding strategy has not changed, but our method of valuing ads based on incrementality has.

## Campaign Incrementality through Time

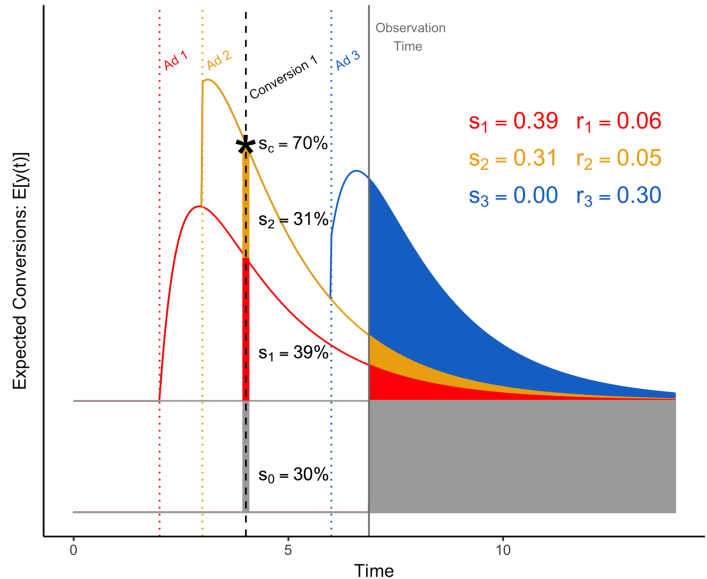

Figure 5 - Partial Incrementality Share, *s*, and Residual Incrementality, *r*

With a clear understanding of both incrementality attribution and bidding, we can now unify the two concepts throughout the duration of the campaign. Above, we defined campaign incrementality three ways:

$$Campaign\ Incrementality = \sum_{i \in users} s_i = \sum_{i \in users} \sum_{j \in impressions} s_{ij} = \sum_{i \in users} \sum_{j \in impressions} \Delta y_{ij}.$$

The last two equalities recommend a simple mid-campaign decomposition:

$$Campaign\ Incrementality = \sum_{i \in users} \sum_{j \in impressions} E[\Delta y_{ij}|t] = \sum_{i \in users} \sum_{j \in impressions} s_{ij}(t) + r_{ij}(t)$$

where, for user *i*'s impression *j* at time *t*, we define $E[\Delta y_{ij}|t] \equiv s_{ij}(t) + r_{ij}(t)$ as the *expected incremental value* at time *t*, $s_{ij}(t) = \sum_c 1(t_j < t_c < t) \cdot s_{ijc}$ as the observable *partial incrementality share*, and $r_{ij}(t) = \int_{t \geq t_j}^{\bar{T}} \Delta y_{ij}(t) dt = \beta w_{ij} \int_{t \geq t_j}^{\bar{T}} f(t - t_j|\theta) dt = \beta w_{ij} \left(1 - F(t - t_j|\theta)\right)$ as the *residual incrementality* or expected value of the residual ad stock, $1 - F(t - t_j|\theta)$. Intuitively, each impression's contribution to a campaign's incrementality is the sum of its incrementality shares from conversions that have happened so far *and* the expected incremental value of the impression's residual ad stock whose incrementality we have yet to observe. For accounting purposes (e.g., to compute performance metrics such as CPIA), we might want to compute each impression's *residual cost*, $r^c{}_{ij}(t) = c_{ij}\left(1 - F(t - t_j|\theta)\right)$, the portion of costs whose incrementality shares and, hence, returns we have not yet observed.

In practice, we might bid on a number of impressions, wait an hour to observe additional conversions, and reestimate the incrementality model to update its coefficients ($\beta$). With an updated model, we would then score all past impressions' partial incrementality share and recompute their residual bids and residual costs. If, however, for computational efficiency, we wish to only score the impressions with nonzero incrementality share, we could instead score each impression's incrementality share as an *expected incrementality share*,
$E[s_{ij}|t] = E[s_{ij}(\tilde{T})|t] = E[s_{ij}(t)|t]/F(t - t_j|\theta) = s_{ij}(t)/F(t - t_j|\theta)$. As the last expression demonstrates, we simply divide the observed partial incrementality share by the impression's realized ad stock to obtain the expected incrementality share.

While these simple formulas do not generalize to the full model, we can compute them using previously computed final quantities, $\beta_{k,\theta}$ and $w_{ijk}$, using $\Delta y_{ij}$ and $r_{ij}(t)$. For the full model, residual incrementality is $r_{ij}(t) = \Sigma_k \Sigma_\theta \beta_{k,\theta} w_{ijk}(1 - F(t - t_j|\theta))$. For expected incrementality shares, $E[s_{ij}|t] = s_{ij}(t) \cdot \frac{\Delta y_{ij}}{\Delta y_{ij} - r_{ij}(t)}$ which multiplies the observed incrementality share by the ratio of the expected incremental value and its difference from the residual incrementality. A similar calculation applies to the residual cost, $r^c{}_{ij}(t) = c_{ij} \cdot \frac{r_{ij}(t)}{\Delta y_{ij}}$.

A viable incrementality platform needs simple metrics that provide an up-to-date forecast of a campaign's incrementality over time. We conclude this section with a glossary of the key concepts, formulas, and definitions. These quantities provide a stable foundation for attribution reporting and "black box" incrementality model training which we discuss in the next section.

| Name | Formula | Definition |
|---|---|---|
| Cost per Exposure | $c_{ij}$ | Cost of impression $j$ shown to user $i$. |
| Ad stock | $A_j = 1(t > t_j)$ | The unit of ad stock generated by impression $j$. |
| Ad stock kernel | $f(t - t_j|\theta)$ | A density, parameterized by $\theta$, that models the distribution of the ad effect over time. |
| Impression Ad Stock | $x_{ij}(t|\theta) = A_j \cdot f(t - t_j|\theta)$ | The distribution of ad stock of shape $\theta$ for impression $j$. |
| Characteristic Ad Stock | $x_{ijk}(t|\theta) = w_{ijk} \cdot x_{ij}(t|\theta)$ | The distribution of ad stock of shape $\theta$ for impression $j$, weighted by characteristic $k$. |
| User Ad Stock | $x_{ik}(t|\theta) = \sum_j x_{ijk}(t|\theta)$ | Level of ad stock for ads with characteristic $k$ shown to user $i$. |
| Baseline Conversions | $\alpha(t|W)$ | The conversion rate in the absence of advertising over time, conditional on the time-varying context $W$. |
| Characteristic Effect | $\Delta y_{ijk} = \sum_\theta \beta_{k,\theta} w_{ijk}$ | The attribution of characteristic $k$ to the incremental impact of impression $j$. |

| Characteristic Effect over Time | $\Delta y_{ijk}(t) = \sum_\theta \beta_{k,\theta} x_{ijk}(t\|\theta)$ | The attribution of characteristic $k$ to the incremental impact of impression $j$ over time. |
|---|---|---|
| Ad Effect | $\Delta y_{ij} = \sum_k \sum_\theta \beta_{k,\theta} w_{ijk} = \sum_k \Delta y_{ijk}$ | The effect of impression $j$ shown to user $i$. |
| Ad Effect over Time | $\Delta y_{ij}(t) = \sum_k \sum_\theta \beta_{k,\theta} x_{ijk}(t\|\theta) = \sum_k \Delta y_{ijk}(t)$ | The effect of impression $j$ shown to user $i$ over time. |
| Incremental Conversions | $\Delta y_i = \sum_j \sum_k \sum_\theta \beta_{k,\theta} w_{ijk} = \sum_j \Delta y_{ij}$ | The total effect of ads on user $i$. |
| Incremental Conversions over Time | $\Delta y_i(t) = \sum_k \sum_\theta \beta_{k,\theta} x_{ik}(t\|\theta) = \sum_j \Delta y_{ij}(t)$ | The total effect of ads on user $i$ over time. |
| Lift per Impression | $\%_j = sign(\Delta y_{ij}) \cdot \frac{\Delta y_{ij}}{\int \alpha(t\|W) \Delta y_{ij}(t) dt}$ | The ad effect relative to baseline conversions, weighted by time-varying incrementality. |
| Conversion Lift | $\%_c = \frac{\Delta y_i(t)}{\alpha(t\|W)}$ | The total effect of ads relative to baseline conversions. |
| Incrementality Share | $s_{ic} = \frac{\Delta y_i(t_c)}{\alpha(t\|W) + \Delta y_i(t_c)}$ | The share of user $i$'s conversion $c$ occurring at time $t_c$ caused by the ads. |
| Cost per Incremental Action | $CPIA = \frac{c_{ij}}{\Delta y_{ij}}$ | Cost of user $i$'s ad exposure divided by its ad effect. |
| Partial Incrementality Share | $s_{ij}(t) = \sum_c 1(t_j < t_c < t) \cdot s_{ijc}$ | Incrementality share accounting for the effect of ad $j$ on conversions up to time $t$. |
| Residual Incrementality | $r_{ij}(t) = \sum_k \sum_\theta \beta_{k,\theta} w_{ijk}(1 - F(t - t_j\|\theta))$ | Expected incrementality that has not yet been observed by time $t$. |
| Expected incremental value at time $t$ | $E[\Delta y_{ij}\|t] \equiv s_{ij}(t) + r_{ij}(t)$ | Forecasted incremental value is the sum of partial incrementality share and residual incrementality. |
| Expected Incrementality Share | $E[s_{ij}\|t] = s_{ij}(t) \frac{\Delta y_{ij}}{\Delta y_{ij} - r_{ij}(t)}$ | Rescaling of observed incrementality shares to forecast expected incremental value. |
| Residual Cost | $r^c_{ij}(t) = c_{ij} \frac{r_{ij}(t)}{\Delta y_{ij}}$ | Portion of costs for which incrementality *has not* been observed by time $t$. |
| Accumulated Cost | $c_{ij}(t) = c_{ij} \frac{\Delta y_{ij} - r_{ij}(t)}{\Delta y_{ij}} = c_{ij} - r^c_{ij}(t)$ | Portion of costs for which incrementality *has already* been observed by time $t$. |
| Expected CPIA | $E[CPIA] = \frac{\sum_{ij \in (Slice)} c_{ij}}{\sum_{ij \in (Slice)} E[\Delta y_{ij}\|t]}$ $E[CPIA\|t] = \frac{\sum_{ij \in (Slice)} c_{ij}(t)}{\sum_{ij \in (Slice)} s_{ij}(t)}$ | Cost per Incremental Action within a given subset "Slice" of users and impressions. |

## "Black Box" Incrementality Model Training

While we would normally want to directly train the incrementality model and use it directly for bidding, many advertisers or even advertising technology companies cannot easily deploy an additional bidding algorithm specific to incrementality. Suppose we have to treat an ad-buying algorithm as a "black box" designed to maximize whatever metric we input. Two examples of black boxes follow:

1. **3rd-Party Algorithmic Buying:** The advertiser can observe enough information to compute an incrementality share on observed conversions, but does not actually do the

  model training. For example, an advertiser could score conversions in real-time and send that information to their marketing partner who runs a programmatic bidder.

2. **Human/Manual Buying:** The advertiser's programmatic advertising employees or agencies hand tune advertising campaigns based on incrementality-powered dashboards.

In both cases we would like to pass information about the expected incremental gains from purchasing advertising to the buyers. As such, to redirect the buyers toward incrementality, we only need to provide the black box our estimates of expected incremental values or expected incrementality shares (of impressions or conversions). This insight can be powerful as many details of the ad buying process, ranging from ad quality evaluations to deep neural network model regularization, are obfuscated from the advertiser's view.

However, in direct contradiction to the thesis of this section, as discussed in Johnson & Lewis (2015), incrementality bidding is a generalization of all other major forms of ad pricing models: cost per impression (CPM), cost per viewable impression (CPV), cost per click (CPC), or cost per action (CPA). In each of those models, we can redefine the outcome metric as a downloaded ad, visible ad, click on the ad, click or view followed by an action, or the desired action. As such, we recommend that, rather than trying to build a separate incrementality testing and scoring system (essentially an incrementality bidder), ad technology partners build an *incrementality compatible* bidder with the aim of running all pricing models, incrementality bidding included, through a single system. In particular, by skipping the black box incrementality model, the details of the ad-buying process that are well-known to the ad buyer can better influence and refine the incrementality estimates and, hence, improve the yield of the end-to-end incrementality-based ad buying.

## Practical Considerations for Incrementality Bidding

With the basics of incrementality bidding and attribution covered, we address two common practical considerations: how can we incorporate smooth time-varying incrementality features and how might using a nonlinear link function affect our ability to estimate the incremental value of an ad? We begin with time-varying features.

### Fourier Series with Dynamic Ad Stock: Exponential and Gamma Distributions

Even in a continuous-time model, time-varying $\alpha(t|W)$ features do not complicate the $\Delta y$ integral. To this end, we can simply include Fourier series terms to model the baseline. However, we would also like to include time-varying conjunction features with ad stock to model whether ads are more incremental at different times of the day or on different days of the week. We propose using Fourier series with exponential-distributed ad stock for user $i$'s impression $j$:

$$x_{ijna}(t|\tau) = A_j w_{ij} \cdot \frac{1}{1-e^{-T/\tau}} \frac{1}{\tau} e^{-(t-t_j)/\tau} \left(a \cdot sin(\tfrac{2\pi n}{S}t) + (1-a) \cdot cos(\tfrac{2\pi n}{S}t)\right).$$

The Fourier series operates on a fixed periodic window of duration $S$, such as a day or week, with $a = 1$ for the sine terms and $a = 0$ for the cosine terms. Intuitively, these features can amplify or dampen the ad effects at different times of day. But most importantly for real-time bidding calculations, the time integrals for these conjunction features between Fourier series and the exponential distribution are analytic. The formulas are especially simple if $S$ divides evenly into $T$, the (optional) truncated upper bound of the exponential model of ad stock:

$$\Delta y_{ij} = \sum_{n=0}^{\{n\}} \sum_{a=0}^{1} \Delta y_{ijna} = \sum_{n=0}^{\{n\}} \sum_{a=0}^{1} \int_{t_j}^{t_j+T} \beta_{na} x_{ijna}(t|\tau) dt$$

$$\Delta y_{ij} = \sum_{n=0}^{\{n\}} \sum_{a=0}^{1} \beta_{na} A_j w_{ij} \frac{1}{1+(\frac{2\pi n}{S}\tau)^2} \left( a \cdot \left( sin(\frac{2\pi n}{S} t_j) + \frac{2\pi n}{S}\tau cos(\frac{2\pi n}{S} t_j) \right) + (1-a) \cdot \left( cos(\frac{2\pi n}{S} t_j) - \frac{2\pi n}{S}\tau sin(\frac{2\pi n}{S} t_j) \right) \right)$$

.

As expected, the integral for $n=0$ is just $\beta_{n=0,a=1} \cdot w_{ij}$, corresponding to the special case where the cosine term of the Fourier series of $x_{ijna}(t)$ simplifies to the exponential distribution.

These exponential features can be generalized to model ad stock using the gamma distribution which has exponential, chi-square, and Gaussian distributions as special cases. Fortunately, the gamma distribution's computation of expected incremental value is [still tractable]:

$$x_{ijna}(t) = A_j w_{ij} \cdot \frac{(t-t_j)^{k-1} e^{-(t-t_j)/\tau}}{\Gamma(k)\tau^k} \left( a \cdot sin(\frac{2\pi n}{S} t) + (1-a) \cdot cos(\frac{2\pi n}{S} t) \right).$$

$$\Delta y_{ij} = \sum_{n=0}^{\{n\}} \sum_{a=0}^{1} \int_{t_j}^{\infty} \beta_{na} x_{ijna}(t) dt = \sum_{n=0}^{\{n\}} \sum_{a=0}^{1} \beta_{na} A_j w_{ij} \cdot \frac{1}{2} \left( a \cdot F_a(t_j|n) + (1-a) \cdot F_{1-a}(t_j|n) \right)$$

where $F_a(t_j|n) = \left( sin(\frac{2\pi n}{S} t_j) - i cos(\frac{2\pi n}{S} t_j) \right) \left( 1 - i \frac{2\pi n}{S}\tau \right)^{-k} + \left( sin(\frac{2\pi n}{S} t_j) + i cos(\frac{2\pi n}{S} t_j) \right) \left( 1 + i \frac{2\pi n}{S}\tau \right)^{-k}$,

$F_{1-a}(t_j|n) = \left( cos(\frac{2\pi n}{S} t_j) - i sin(\frac{2\pi n}{S} t_j) \right) \left( 1 + i \frac{2\pi n}{S}\tau \right)^{-k} + \left( cos(\frac{2\pi n}{S} t_j) + i sin(\frac{2\pi n}{S} t_j) \right) \left( 1 - i \frac{2\pi n}{S}\tau \right)^{-k}$, and

$i = \sqrt{-1}$ in the complex plane, though $F_a(t_j|n)$ and $F_{1-a}(t_j|n)$ are both are real-valued functions for $t_j \geq 0$. Further simplifications exist for $k \in \{1, 2, 3, ...\}$. Computing residual incrementality, however, is more [numerically challenging], though there are efficient computational methods for estimating the upper incomplete gamma function, $\Gamma(k, x) = \int_{x}^{\infty} t^{a-1} e^{-t} dt$, used below:

$$r_{ij}(t) = \int_{t \geq t_j}^{\infty} \Delta y_{ij}(s) ds = \sum_{n=0}^{\{n\}} \sum_{a=0}^{1} \beta_{na} \int_{t \geq t_j}^{\infty} x_{ijna}(s) ds = \sum_{n=0}^{\{n\}} \sum_{a=0}^{1} \beta_{na} A_j w_{ij} \cdot \frac{1}{2} \left( a \cdot F_a(t|t_j, n) + (1-a) \cdot F_{1-a}(t|t_j, n) \right)$$

$$F_a(t|t_j, n) = \left( sin(\frac{2\pi n}{S} t_j) - i cos(\frac{2\pi n}{S} t_j) \right) \left( 1 - i \frac{2\pi n}{S}\tau \right)^{-k} \cdot \Gamma\left( k, \left(\frac{t-t_j}{\tau}\right) \left( 1 - i \frac{2\pi n}{S}\tau \right) \right) / \Gamma(k) +$$
$$\left( sin(\frac{2\pi n}{S} t_j) + i cos(\frac{2\pi n}{S} t_j) \right) \left( 1 + i \frac{2\pi n}{S}\tau \right)^{-k} \cdot \Gamma\left( k, \left(\frac{t-t_j}{\tau}\right) \left( 1 + i \frac{2\pi n}{S}\tau \right) \right) / \Gamma(k)$$

$$F_{1-a}(t|t_j, n) = \left( cos(\frac{2\pi n}{S} t_j) - i sin(\frac{2\pi n}{S} t_j) \right) \left( 1 + i \frac{2\pi n}{S}\tau \right)^{-k} \cdot \Gamma\left( k, \left(\frac{t-t_j}{\tau}\right) \left( 1 + i \frac{2\pi n}{S}\tau \right) \right) / \Gamma(k) +$$
$$\left( cos(\frac{2\pi n}{S} t_j) + i sin(\frac{2\pi n}{S} t_j) \right) \left( 1 - i \frac{2\pi n}{S}\tau \right)^{-k} \cdot \Gamma\left( k, \left(\frac{t-t_j}{\tau}\right) \left( 1 - i \frac{2\pi n}{S}\tau \right) \right) / \Gamma(k).$$

## Computing Incremental Value using Non-Additive Functions of Ad Stock

For a non-additive, differentiable continuous-time function $y_{ij}(t)$, we can use three approaches to estimate the incremental value of an ad impression. The first approach directly computes the difference in the integral of $y_i(t)$ over time. The second and third approaches instead use the derivative of $y_i(t)$ with respect to $A_j$ (omitting subscripts and recentering $t_j = 0$ for clarity) and either evaluate the derivative at $A = 1$ or integrate with respect to $A$ on $[0, 1]$. Leibniz Rule allows the differentiation of $y_i(t)$ under the time integrals.

1. Integrate after evaluating $y(A, t)$ at $A \in \{0, 1\}$: $\Delta y = \int_0^{\bar{T}} y(A = 1, t) dt - \int_0^{\bar{T}} y(A = 0, t) dt$

2. Evaluate at $A = 1$: $\Delta y = \frac{\partial}{\partial A} \int_0^{\bar{T}} y(A, t) dt = \int_0^{\bar{T}} \frac{\partial y}{\partial A}(A, t) dt$

3. Integrate $A \in [0, 1]$: $\Delta y = \int_0^1 \frac{\partial}{\partial A} \left( \int_0^{\bar{T}} y(A, t) dt \right) dA = \int_0^1 \left( \int_0^{\bar{T}} \frac{\partial y}{\partial A}(A, t) dt \right) dA$

Since all three methods require an integral over time, computing incremental value in nonlinear models can be quite challenging. For this reason, we opt for the linear model which is easier to estimate and trivial to integrate over time. However, computing incrementality share does not require this time integral---potentially facilitating practical nonlinear attribution models via "black box" incrementality model training.

# Continuous-Time Causal Training

We have described the motivations for a continuous-time model to enable precise calculations of incrementality share, incremental values, and their implied incrementality-based bids. With that foundation we now present several practical considerations that arise in industrial-scale modeling and training.

## Computational Tractability: Down-Sampling Negatives in Continuous Time

We return to our simple continuous-time model:

$$y_i(t) = \alpha(t) + \beta x_i(t) + \varepsilon_i(t).$$

In this model, the outcome (conversions) arrive at specific moments in time---a set of measure zero: $y_i(t) = 0$ except at $t = t_c$ for users who convert. Normally, in a continuous time model, we might consider a full integral over the entire set of time during which we observe user $i$'s behavior. However, in this case we are going to leverage the fact that conversions are sparse to approximate the continuous time integral, but only up to a practical degree of approximation. In a companion paper (see Appendix), Lewis (2018) provides a more complete discussion of the method and motivation. Here we simply apply the method.

**Sampling Procedure:**
1. *Positives* {+}: Collect all (users with) conversions.
2. *Negatives* {-}: Collect a random sample of all users (both with & without a conversion).
3. For each {+}: Evaluate the model's regressors, $\alpha(t)$ and $x_i(t)$, at $t = t_c$, the time of the conversion.

a. *"Double-Negatives" {+} → {+0}*: Augment the data with $\alpha(t_c)$ and $x_i(t_c)$ and $y(t_c) = 0$. Not required for outcomes which are sparse with respect to the measure: $\#\{+\} << NT$.

4. For each {-}: Evaluate the model's regressors, $\alpha(t)$ and $x_i(t)$, at $t = t_r$, a random time.

**Training Procedure:**

1. Set the sample weights for the model to $w^- = \frac{NT}{\#\{-\}}$, $w^+ = 1$, and $w^{+0} = -1$.

   $NT = \sum_{i \in Users} t_{i,end} - t_{i,start}$ represents the total measure of $N$ users over $T$ units of time (e.g., days) of sampling.

2. Train the model.

For models with non-sparse outcomes, we risk sample-selection bias from unconditionally sampling the positives and uniformly sampling the negatives. In short, we have double-counted the number of observations present in the positives' part of the distribution of covariates. In some models, such as ordinary least squares, this can be resolved by simply setting $w^+$ to a large number such as $w^+ = \frac{1}{N}$ and then rescaling the final coefficient estimates by $N$. This works by letting the denominator of $\hat{\beta}_{OLS} = (X'WX)^{-1}(X'WY)$ be primarily determined by the negatives---the set of full measure---while the numerator is determined by set of positives. More generally, we can remove this continuous-time sampling bias from any weighted estimator by appending a negatively weighted observation for each positive: $< Y = 0, X = x_i(t_c), w^{+0} = -1 >$. We call these observations "double negatives" due to $Y = 0$ (e.g., a "negative" in machine learning parlance) and $w^{+0} = -1$ (e.g., a negative observation in terms of weight).

Conveniently, by building the training data, we have built the features necessary to compute incrementality shares for conversions, impressions, and bids. After estimating the model, we will have the model coefficients. Further, because we would like to use all of the positives to ensure the most precise ad effectiveness estimates possible, we can score all of our conversions, impressions, and even estimate incrementality opportunities using the negatives' representative sample of impressions.

But how much model precision did we potentially forego by only utilizing a sample of negatives? Lewis (2018) argues that, for a simple binomial model in which we sample $C = \frac{\#\{-\}}{\#\{+\}}$ times as many negatives as positives, down-sampling of the negatives produces estimates whose variance is 1+1/C times larger than the theoretical lower bound if there were, in fact, an infinite number of independent negatives that we could draw from. In practice, choosing $C = 10$ ensures that at most 10% of available precision remains untapped; however, additional reductions in variance are quadratic in computational costs of sampling $N$ because the variance is $O(1)$. This is worse than the usual $O(1/N)$ because we are unable to sample more positives with our increase in negative samples. In short, by constructing our sample in continuous time by down-sampling, we both avoid reducing the precision of our time scale and effectively eliminate any additional computational burden of traditional integrated continuous-time estimation of panel data.

## Continuous-Time Endogeneity

Why would we want to estimate a continuous-time model rather than a more standard batch aggregation model (e.g., hourly, daily, weekly)? We provide three reasons.

First, a continuous-time model provides tractable time-variability of baseline conversion rates and ad effects.

Second, it provides time-scale flexibility for ad effects. Depending on the advertiser and the conversion outcome, an ad effect's relevant window can vary on the scale of seconds, minutes, hours, or days. Ad technology companies need a system that is robust to wide variability across advertisers' outcomes' relevant time scales where training batches are as free as possible from constraints of arbitrary time scales. For example, if some advertisers define conversions that are similar or equivalent to ad exposure, ad viewing, or clicking, time delays will be very short between the bid and conversion events. On the other hand, some advertisers could define conversion events such as booking travel or visiting a brick-and-mortar store for which ad effects may take multiple days or weeks to be fully realized.

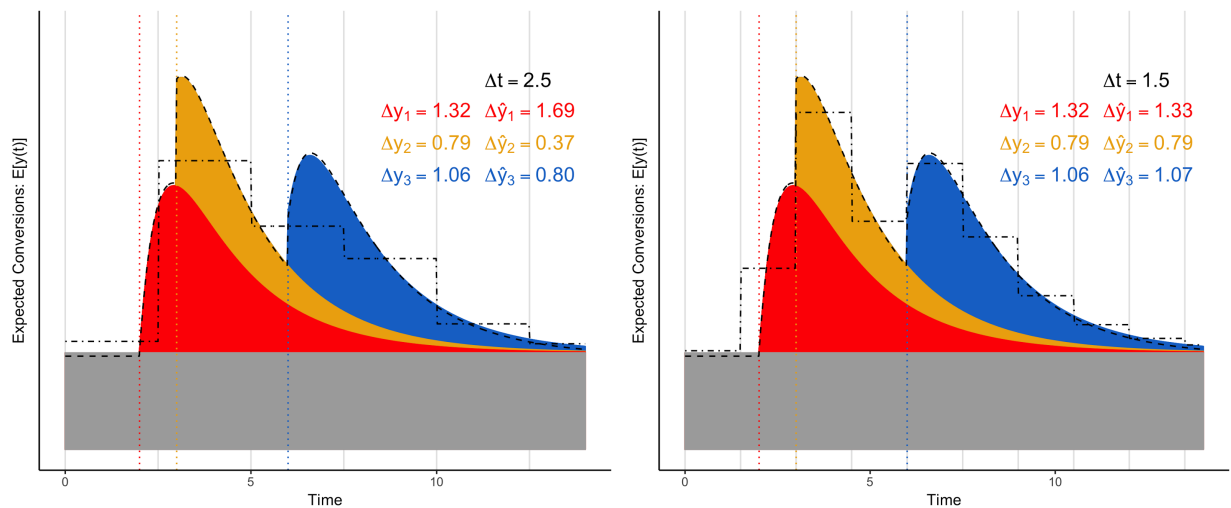

Figure 6 - Bias Emerges if Panel Sampling Rate is Slower than Targeting Feedback

The third reason is perhaps the most important for estimating incrementality: the microsecond time scale of endogeneity in real-time bidding and ad-targeting systems. This endogeneity emerges due to common constructs such as positive and negative targeting and frequency capping. Ad targeting models update their bidding behavior in real-time, introducing endogeneity (feedback) into the training data and biasing incrementality estimates if the time scale of the feedback is faster than our training data features' temporal precision (e.g., hourly). For example, if ads cause a purchase which leads the bidder to negatively target (i.e., stop showing ads), we will induce a negative correlation between ad exposure and purchases---not because ads reduce purchases but rather because of reverse causality: purchases cause a reduction in ads! Figure 6 illustrates the presence and absence of bias in three impressions' estimates of $\Delta y$ when the sampling rate goes from $\Delta t = 2.5$ to $\Delta t = 1.5$ in the presence of negative targeting. Continuous-time modeling can avoid this source of bias while still accommodating complex targeting features.

## Retargeting Features in Continuous Time

Continuous-time modeling ensures that the information set assumed during training is the same as the information set when bidding. For example, suppose we accidentally include a feature in our model such as "number of impressions seen after conversion time" explicitly or one that is implicitly correlated such as "number of impressions seen today." Then, if we negatively target converters, we implicitly include the conversion as a predictor and bias our pricing downward because people who convert will see fewer ads after converting, introducing a mechanical negative correlation. Continuous-time modeling prevents this problem.

We can include retargeting-compatible features for any event such as "user visited the homepage within the last X minutes/hours/days." We can generate these as binary, counts, splines, categories, or even dynamic "event stock" features using outcome events along the purchase funnel: homepage, product pages, and purchase confirmation. We can also generate analogous features using past ad exposures or clicks. If these events enhance the effectiveness of additional ad impressions, the model will increase the ad's valuation accordingly. If, however, these events suggest a decrease in ad effectiveness, the model will decrease the ad's expected value, perhaps to zero---implying model-driven negative targeting and frequency capping. As long as these features are constructed based on moving windows of events, any increase or decrease in the expected value of an additional ad will be transitory with respect to that moving window. "Event stock" features let the model automatically estimate the relevant time horizon.

We can construct retargeting "event stock" features by dynamically weighting the ad stock using the set of retargeting events, $R$, $w_{ij}(t|R) = \sum_{r \in R} A_{jr} x_{ir}(t)$. These features represent a double sum over the retargeting events and ads purchased after the event: $A_{jr} = 1(t > t_j \ \& \ t > t_r \ \& \ t_j > t_r)$.

$$x_{ijr}(t|\theta, \theta_R) = x_{ij}(t|\theta) \cdot x_{ir}(t|\theta_R) = A_j f\left(t - t_j | \theta\right) \cdot A_r f_R\left(t - t_r | \theta_R\right)$$

$$x_{iR}(t|\theta, \theta_R) = \sum_{j \in impressions} \sum_{r \in R} A_{jr} x_{ijr}(t|\theta, \theta_R).$$

We provide a simple analytic example here using the exponential distribution for both retargeting event stock and ad stock for the dynamic event-weighted ad stock:

$$x_{ij}(t) = A_j \frac{1}{\tau} e^{-(t - t_j)/\tau}$$

$$x_{ir}(t) = A_r \frac{1}{\tau_R} e^{-(t - t_r)/\tau_R}$$

$$x_{ijr}(t) = A_j \frac{1}{\tau} e^{-(t-t_j)/\tau} \cdot A_r \frac{1}{\tau_R} e^{-(t-t_r)/\tau_R} = A_j A_r \frac{1}{\tau \tau_R} e^{-\left(t\cdot(1/\tau + 1/\tau_R) - (t_j/\tau + t_r/\tau_R)\right)}$$

$$\Delta y_{ijr} = \int_{t_j}^{\infty} \beta_r x_{ijr}(t) dt = \beta_r \sum_{r \in R} A_{jr} \frac{1}{\tau + \tau_R} e^{-(t_j - t_r)/\tau_R}$$

Because $x_{ijr}(t)$ is just a scaled exponential distribution, our other results about <u>Fourier series and exponentials</u> apply after first factoring out $\frac{1}{\tau + \tau_R} e^{-(t_j - t_r)/\tau_R}$ and then using $\bar{\tau} = \frac{\tau \tau_R}{\tau + \tau_R}$.

## Efficient Instrumental Variables Modeling

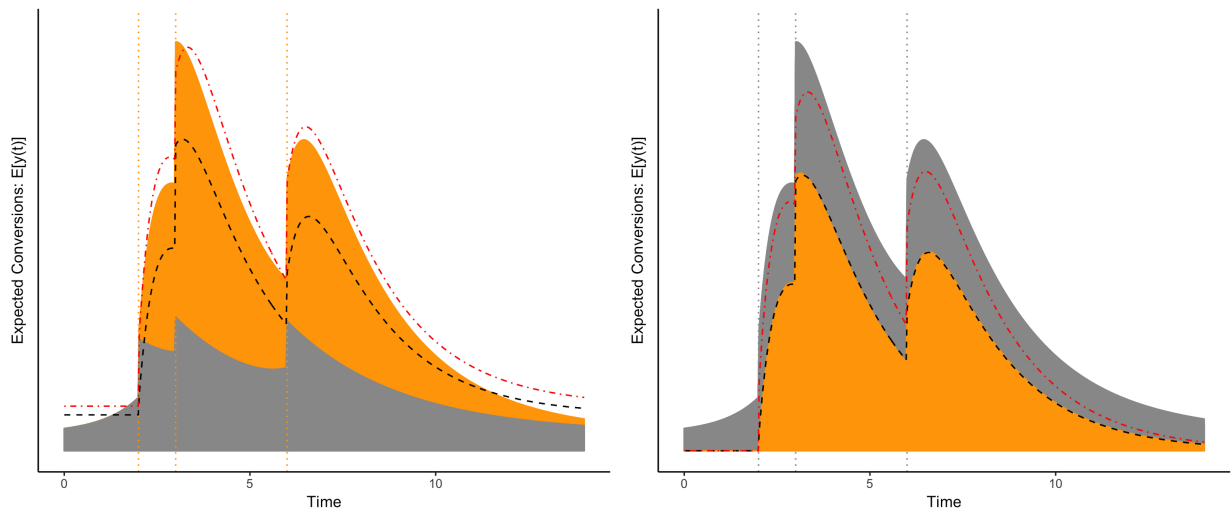

Figure 7 - IV (black dash line) fits worse than OLS (red dot-dash line), but removes endogeneity bias from the incrementality estimates

In <u>Estimating Incrementality</u>, we discussed various sources of endogeneity common to ad effectiveness applications and how creating instrumental variables through randomization can eliminate such bias. Figure 7 illustrates the two points: IV is noisier than correlational methods such as OLS, but provides consistent estimates of the causal parameters used to compute incrementality. Taking the IV's consistency as given, we now discuss how to minimize the drawback of IV's extra noise.

The use of instrumental variables to disregard the portion of variance in treatment (ad exposure) that might be endogenous implies that causal models are going to be inherently noisier. In fact, there is no guarantee that an IV model has an $R^2 > 0$. Negative $R^2$ can happen when the endogeneity is larger than the predictive power of the instrumental variable; however, this is not a reflection of the quality of the model but rather a reflection of causal modeling being inherently noisier than correlative modeling.

Naturally, we would like to improve the precision of our causal estimates. This can be done two ways:

1. Reduce the variation in the second-stage residual, $\varepsilon_i$, by including other covariates ($W$) that are both exogenous ($E[W'\varepsilon] = 0$) and predictive ($E[\beta_W] \neq 0$). This reduces the overall model uncertainty.
2. Predict away the variation in the first-stage residual, $\nu_i$, either through other useful second-stage covariates $W$ or additional instrumental variables $Z$. The former reduces the magnitude of potential endogeneity while the latter increases the strength of our causal model, by either explicitly (an additional randomization) or implicitly (improved modeling) adding additional exploration data.

Adding additional covariates $W$ affects the regression model in the usual ways; however, adding additional instrumental variables $Z$ can present pitfalls. When the number of instrumental variables (e.g., randomized features) equals the number of endogenous variables (e.g., ad exposure), we call the model *exactly identified*. This means that the instrumental variables provide enough linearly independent variation to estimate the model. If the number of instruments was less than the number of endogenous variables, the model would be *under identified*. However, to improve power, we are interested in including additional instrumental variables either by running multiple experiments or by including nonlinear transformations of the instrumental variables (assuming the instrument is more than just a treatment/control binary variable) such as polynomials ($Z^2, Z^3, \dots$) or splines ($(Z - \zeta_1) \cdot 1(Z > \zeta_1), (Z - \zeta_2) \cdot 1(Z > \zeta_2), \dots$). Such a model would be *over identified* by having more instruments than endogenous variables.

The terminology of over/exact/under identification derives from the fact that the instrumental variables model implies solving a system of equations $g(Z)'\varepsilon(\beta) = 0$ for our second stage's regression coefficients, $\beta$. When there are more equations than unknown parameters, the functional form of $g(Z)$ matters for the variance of our linear model's estimates: $Var(\hat{\beta}_{IV}) \propto (X'g(Z)g(Z)'X)^{-1}$. As such, we would like $g(Z)$ to maximize $Corr(g(Z), X)$ ---in short, we want $g(Z)$ to be the best prediction of $X$ possible. Hence, the choice of $Z$ and $g(Z)$ should reflect the structure of their causal impact on $X$ in order to efficiently make use of the exogenous variation in the instrument $Z$ being used to identify the causal effect of $X$. We discuss how to construct such efficient instruments for our continuous-time model in the next section.

## Efficient Instrumental Variables in RTB using Predicted Ghost Ads

Suppose we seek to model the causal effect of advertising impressions in a real-time bidding (RTB) environment where the advertiser who wins an ad impression is, without loss of generality, determined by a second-price auction. Recall the form of the features in our continuous-time ad stock model:

$$x_i(t) = \sum_{j \in impressions} w_{ij} \cdot x_{ij}(t)$$

$$x_{ij}(t) = A_j \cdot f\left(t - t_j | \theta\right).$$

We define a relatively weak instrumental variable as a simple transformation of this feature:

$$z_i(t) = \sum_{j \in bids} w_{ij} \cdot z_{ij}(t)$$

$$z_{ij}(t) = B_j \cdot f\left(t - t_j | \theta\right).$$

Here, we define $B_j$ as bid event $j$, in contrast to $A_j$'s definition as ad impression $j$. In particular, $B_j$ is defined by whether we randomly did or did not submit our bid to the ad exchange for consideration in the auction. As such, $z_{ij}(t)$ is proportional to the *potential* ad stock from winning the impression. However, $z_i(t)$ is *not*, in general, proportional to the potential ad stock from winning because the probability of winning the auction can be very heterogeneous across each bid opportunity $j$ and user $i$. This implies a simple adjustment to the model to unambiguously improve its predictive power, similar to "predicted Ghost Ads" (Johnson, Lewis, and Nubbemeyer 2017), by using a predicted win probability for each auction, $Pr(b_j \ wins)$, given the bid we submitted $b_j$:

$$z_{ij}(t) = B_j \cdot Pr(b_j \ wins) \cdot f\left(t - t_j | \theta\right).$$

In practice, RTB impression volume is sufficiently large that $Pr(b_j \ wins)$ can generally be estimated much more precisely than the conversion rate and, hence, its estimation uncertainty can be disregarded. In fact, for many RTB demand-side platforms (DSPs) who represent multiple advertisers, the clearing price for a given bid request can be known with certainty if another advertiser whom the DSP represents wins the final ad auction or even just an internal auction within the DSP. In such cases, $Pr(b_j \ wins) = 0$. This information improves the statistical power of $z_i(t)$ at predicting $x_i(t)$. However, even though $B_j$ is exogenous, $b_j$ is potentially endogenous because winning or losing the auction could be correlated with the outcome. In short, unless $b_j$ is independent of the outcome (random noncompliance or unconfoundedness), we have reintroduced endogeneity. We resolve this with "ghost bid stock."

In order to restore conditional independence, we construct the analogous exogenous $W$ regressors via "ghost bids" or "predicted ghost ads" (Johnson, Lewis, and Nubbemeyer 2017). Here, these continuous-time "ghost bid stock" are defined based on the user's context $W_i(t)$; we denote them by $\xi$:

$$\xi_i(t) = \sum_{j \in bids} w_{ij} \cdot \xi_{ij}(t)$$

$$\xi_{ij}(t) = G_j \cdot f\left(t - t_j | \theta\right) \text{ or } \xi_{ij}(t) = G_j \cdot Pr(g_j \ wins) \cdot f\left(t - t_j | \theta\right)$$

where $G_j$ is defined by whether we would have submitted a bid if the user were in the treatment group, regardless of random assignment, and $g_j$ is the actual bid that we wanted to submit. The former reflects using "ghost bids" and the latter reflects using "predicted ghost ads," $Pr(g_j \ wins)$, to estimate the total potential ad stock available to user $i$ at time $t$. These features can improve the statistical power of the instrumental variables by better predicting realized ad stock $x_i(t)$ based on whether the user has low or high potential ad stock $\xi_i(t)$.

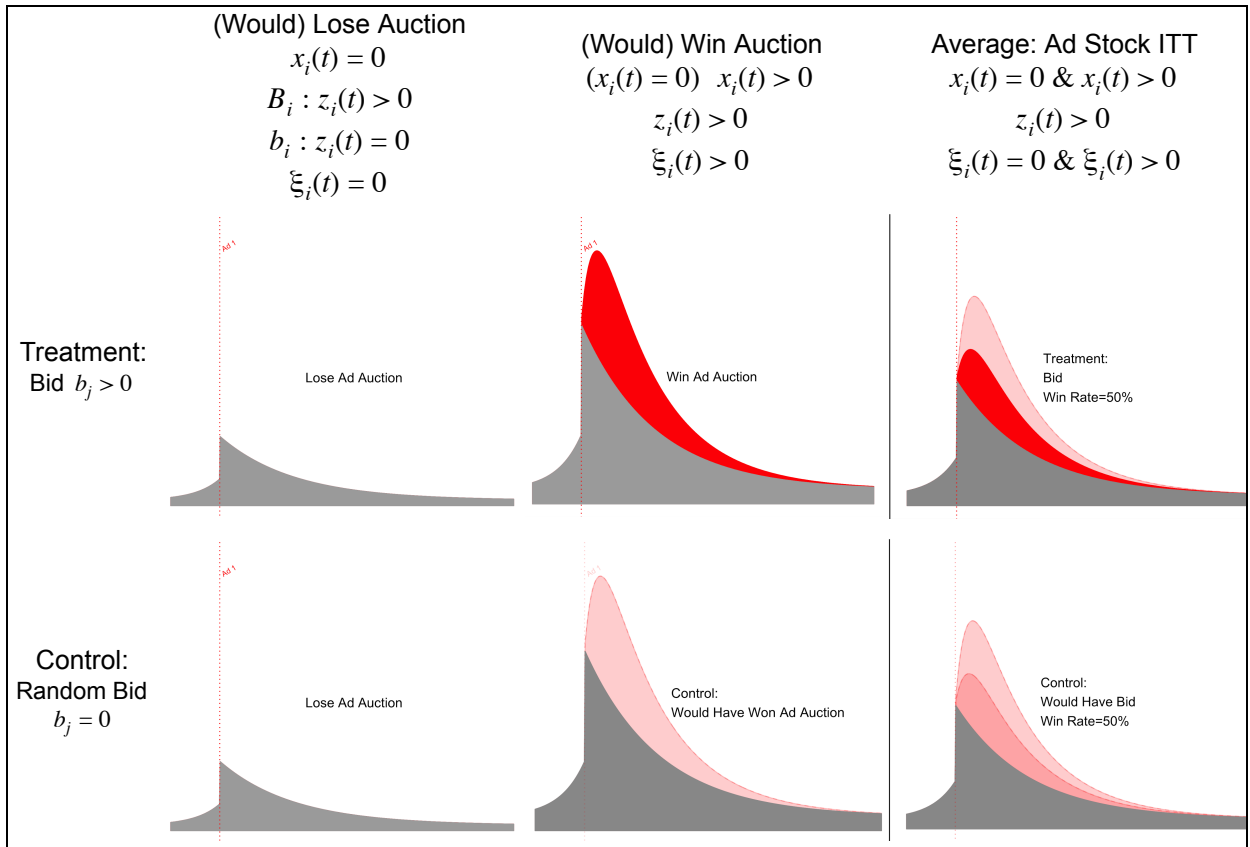

Figure 8 - Nonrandom noncompliance causes the baseline to vary by exposure; causal effect of ad stock can still be recovered using potential ad stock and ghost bid stock.

| Name | Formula | Definition |
|------|---------|------------|
| Ad Stock Kernel | $f(t - t_j \mid \theta)$ | A density, parameterized by $\theta$, that models the distribution of the ad effect over time. |
| Ad Stock | $A_j = 1(t > t_j)$ | The unit of ad stock generated by impression $j$. |
| Impression Ad Stock | $x_{ij}(t \mid \theta) = A_j \cdot f(t - t_j \mid \theta)$ | The distribution of ad stock of shape $\theta$ for impression $j$. |
| Characteristic Ad Stock | $x_{ijk}(t \mid \theta) = w_{ijk} \cdot x_{ij}(t \mid \theta)$ | The distribution of ad stock of shape $\theta$ for impression $j$, weighted by characteristic $k$. |
| User Ad Stock | $x_{ik}(t \mid \theta) = \sum_j x_{ijk}(t \mid \theta)$ | Level of ad stock for ads with characteristic $k$ shown to user $i$. |
| Bid | $b_j$ | The bid submitted for bid opportunity $j$. |
| Bid Stock | $B_j = 1(t > t_j)$ | Randomized intention to submit bid $j$. |
| Potential Ad Stock (simplified) | $z_{ij}(t \mid \theta) = B_j \cdot f(t - t_j \mid \theta)$ | The distribution of potential ad stock of shape $\theta$ for bid $j$. |
| Potential Ad Stock | $z_{ij}(t \mid \theta) = B_j \cdot Pr(b_j\ wins) \cdot f(t - t_j \mid \theta)$ | The distribution of potential ad stock of shape $\theta$ for bid $j$. |
| Characteristic Potential Ad Stock | $z_{ijk}(t|\theta) = w_{ijk} \cdot z_{ij}(t|\theta)$ | The distribution of potential ad stock of shape $\theta$ for bid $j$, weighted by characteristic $k$. |
| User Potential Ad Stock | $z_{ik}(t|\theta) = \sum_j z_{ijk}(t|\theta)$ | Level of potential ad stock for bids with characteristic $k$ shown to user $i$. |
| Ghost Bid | $g_j$ | The bid we want to submit for impression $j$; we randomly perturb this bid as $b_j$ for causal identification. |
| Ghost Bid Stock | $G_j = 1(t > t_j)$ | The unit of bid stock from our intention to submit bid $j$. |
| Ghost Bid Stock (simplified) | $\xi_{ij}(t|\theta) = G_j \cdot f(t - t_j|\theta)$ | The distribution of ghost bid stock of shape $\theta$ for impression $j$. |
| Ghost Bid Stock | $\xi_{ij}(t|\theta) = G_j \cdot Pr(g_j \ wins) \cdot f(t - t_j|\theta)$ | The distribution of ghost bid stock of shape $\theta$ for impression $j$. |
| Characteristic Ghost Bid Stock | $\xi_{ijk}(t|\theta) = w_{ijk} \cdot \xi_{ij}(t|\theta)$ | The distribution of ghost bid stock of shape $\theta$ for impression $j$, weighted by characteristic $k$. |
| User Ghost Bid Stock | $\xi_{ik}(t|\theta) = \sum_j \xi_{ijk}(t|\theta)$ | Level of ghost bid stock for ads with characteristic $k$ shown to user $i$. |

## IVs for Ad Viewability and Other Ex-Post Observable Ad Interactions

For the sake of simplicity, we have avoided the topic of ad viewability. However, bids and viewable ads have the same relationship as bids and impressions: you usually only learn a given impression's viewability, $V_j = \{1 \ if \ viewable, \ 0 \ otherwise\}$, if you win the auction and show the ad. This could lead us to construct compound features and instruments for this case:

$$x_j(t) = A_j \cdot V_j \cdot f\left(t - t_j|\theta\right),$$

$$z_{ij}(t) = Pr(b_j \ wins) \cdot Pr(j \ viewable \ | \ win) \cdot f\left(t - t_j|\theta\right),$$

$$\text{and } \xi_{ij}(t) = Pr(g_j \ wins) \cdot Pr(j \ viewable \ | \ win) \cdot f\left(t - t_j|\theta\right).$$

As with $Pr(b_j \ wins)$ modeling, $Pr(j \ viewable \ | \ win)$ modeling can be done much more precisely than the conversion rate, and its estimation uncertainty can usually be disregarded. Many similar features can be constructed using viewability, duration, clicks, and other endogenous outcomes of the ad interaction experience that are known ex-post.

However, in order to formulate bids using such endogenous ex-post observable features, we would need to replace the model's "viewable impression" feature, which is unobservable at bid request time, with its "expected viewability" weight $w_{ij} = E[V_j] = Pr(j \ viewable \ | \ win)$. By so doing, we transform the ex-post observable "viewable impression" features into ex-ante predictable "impression" units that we are bidding for in the auction. Similar transformations from ex-post observable attribution features to ex-ante predictable features preserve harmony between further improvements to the attribution model using ex-post observable characteristics while maintaining feature set parity with the ex-ante predictions required for bidding.

## User- versus Bid-Level Randomization: Tradeoffs

The choice between user- and bid-level randomization affects several aspects of incrementality: consistency of causal estimation, bias in ghost bids, statistical power, and easy A/B test reporting.

We begin with two major issues with ghost bids: bias can be a serious concern with user-level randomization by either failing to identify the causal effect of marginal ad stock or by invalidating the instrumental variables as the distribution of ghost bids diverge over time between treatment and control users.

While ghost bid stock features are not required for consistent causal estimation under bid-level randomization, these features are critical under user-level randomization. We first define bid-level randomization as assigning a bid to the treatment group with probability $p$ using a random draw from the uniform distribution $U_{ij}$ on $[0,1]$ for bid $j$, $b_j = g_j \cdot 1(U_{ij}[0,1] < p)$ ; user-level randomization is done analogously, but where the uniform random draw, $U_i$, only varies across users but not across bids for a given user. Under user-level randomization, the total accrued ad stock $x_i(t)$ is a simple composition of ghost bid stock $\xi_i(t)$ available to user $i$ and the user-level randomization: $x_i(t) = \xi_i(t) \cdot 1(U_i[0,1] < p)$ . This implies that we must include $\xi_i(t)$ as a control variable in order to obtain an unbiased estimate of the average causal effect of increasing ad stock. Otherwise, if users who accrue more ad stock convert at a higher (or lower) rate than average, the average causal effect of increasing ad stock (e.g., showing ads) will be upward (or downward) biased. More generally, under user-level randomization and without $\xi_i(t)$ , we cannot measure the causal effects of advertising heterogeneity: ad frequency, size, placement, creative, viewability, etc. As a result, our valid instrumental variables significantly weaken to simply $z_i(t) = 1(U_i[0,1] < p)$ , the user-level random assignment.

However, in a user-state-dependent bidding algorithm, user-level randomization causes a potential violation of the required assumption of exogeneity of $\xi_i(t)$ : $E[g_j|Treatment] \neq E[g_j|Control] \ \forall j > 1$ . Upon submitting our first bid, we can potentially treat the user differently by showing the ad or not. Then, exogeneity of $\xi_i(t)$ is violated for subsequent bids if our bidding algorithm depends on past exposures, either explicitly via frequency capping or implicitly via changes in any user behavior caused by the advertising that changes future model bids (e.g., a form of endogeneity referred to as "covariate shift" in the machine learning literature). Johnson, Lewis, and Nubbemeyer (2017) provide examples of this bias in their implementation of predicted ghost ads. This weakness of predicted ghost ads in user-level randomization should discourage the use of user-level randomization as the primary source of exogeneity for estimating statistically precise incrementality bidding or attribution models.

Statistical power depends on the decay rate of ad stock versus the delay between bid opportunities. If there are many bid opportunities between exposure and conversion, bid-level randomization will result in every user having ghost bid ad stock close to the average, leading to low statistical power. On the other hand, if incremental conversion delay is very short, then bid-level randomization suffers no loss in power.

Finally, A/B-style incrementality test reporting is straightforward with user-level randomization and can be made more statistically precise with ghost bids or predicted ghost ads: collect all users in the treatment group who are eligible for ad exposure and compare them to the users who would have been eligible in the control group. In contrast, bid-level randomization requires an incrementality share scoring model and answers a slightly different counterfactual question: "How incremental are my ads on the average user?" versus "How incremental is my average impression?" These are usually the same question, except for specific modeling nuances. However, the event-based modeling extends the precision gains from ghost bids or predicted ghost ads by going beyond user-level eligibility to explicitly modeling each ad exposure.

While hybrid approaches to bid- and user-level randomization could balance some of these tradeoffs, we believe that bid-level randomization generally provides the simplest, most robust, and easiest to maintain incrementality bidding architecture and can be especially powerful when coupled with an efficient explore-exploit strategy which we discuss in [Bayesian Bootstrap for Confidence Intervals & Exploration](#).

## Surrogates

Usually, the most business-relevant end-of-funnel outcomes such as profits are simply too noisy to use for ad effectiveness evaluation (Lewis & Rao 2015). As such, others (e.g., Athey et al. 2016) have proposed using proxies or *surrogates* of the outcome of interest. For example, if effective ads must influence upper funnel activity by causing incremental visitors or visits, we could instead model incrementality of those outcomes instead of profits and then construct a model to convert incremental visitors to incremental profits (see Johnson, Lewis, & Nubbemeyer 2017).

Using upper-funnel proxies improves the statistical power of our continuous-time model: more outcomes with larger lifts in closer proximity to the advertising. More outcomes following the ads increases precision. Larger lifts makes the effects easier to detect. And closer proximity of the incrementality reduces the background noise by the model implicitly ignoring many delayed conversions that are unlikely to be incremental.

Deciding to use a surrogate metric rather than the metric of interest represents yet another example of the classical bias-variance tradeoff. For example, if we are trying to maximize profits but use a lower variance, upper funnel metric such as visits to the website as a surrogate, we know that, by virtue of not directly optimizing for the right metric that we *may* suffer bias. For economically calibrated decision-making, we could model incremental profits (the consistent model) in terms of website visits (the efficient model) and then applying Hausman penalization (Lewis & Wong 2018), described in the next section, to the predicted incrementality in profits as predicted based on the incrementality in website visits. Hausman penalization will let us use the information in profits to "causally correct" some coefficients of the more precise incrementality predictions based on website visits where it is making statistically detectable mistakes without passing through all of the noise of profits. In short, if the estimated incrementality based on profits is statistically indistinguishable from the estimated incremental profits based on website visits, we might simply prefer the more precise model to prevent too erratic bidding behavior.

# Production-Ready Causal Machine Learning

In this section we discuss a production-ready causal machine learning system. We define its requirements and our estimator of choice that satisfies its requirements.

## Requirements: Causal, Predictive, Scalable, and Efficient

A production-ready causal machine learning model must have the following properties:
- **Causal**: Its predictions are not dependent on the distribution of the training data remaining stable. For example, if we train a model offline, its online performance should be very similar, if not identical to the offline performance.
- **Predictive**: Its predictions are as precise as possible out of sample. In short, the model accommodates model "regularization" tuning via a valid, automatic, and feasible cross-validation procedure.
- **Scalable**: The model can be estimated with a large number of sparse features---its computational complexity is approximately linear in the number of features or data sparsity. In practice, this means "can be estimated without computing a feature matrix inverse." Acceptable estimation procedures include preconditioned conjugate gradient (PCG), limited memory Broyden–Fletcher–Goldfarb–Shanno (L-BFGS), and stochastic gradient descent (SGD) algorithms in increasing order of scalability.
- **Efficient**: Within the relevant class of similar estimators, it is the minimum variance estimator. In practice, this usually means that there does not exist a reweighting of the data that will have smaller confidence intervals while still estimating the same quantity.

With the end of designing our ideal estimator, we will initially focus on the linear regression model for the sake of pedagogy.

## Optimal Instrumental Variables Estimator: Causal & Efficient

We begin our ideal estimator by identifying the most efficient causal IV estimator. The linear IV model is our starting point as a consistent causal estimator: $plim_{N \to \infty} \hat{\beta}_{IV} = \beta$. A statistically efficient causal estimator minimizes the variance among the class of causal estimators. Two-stage least squares (2SLS) is the minimum variance linear estimator among the class of linear IV estimators. However, the minimum variance IV estimator would minimize $Var(\hat{\beta}_{IV}) \propto \sigma^2 (X'g(Z)g(Z)'X)^{-1}$ which would employ a potentially nonlinear function $g(\cdot)$ to maximize the covariance between $X$ and $Z$ and $Y$ and $Z$ to maximize the denominator, $(X'g(Z)g(Z)'X)^{-1}$, and minimize the residual variance, $\sigma^2$. As such, we can use nonlinear basis functions of $Z$ in conjunction with nonlinear regularization to estimate these complex relationships.

## Hausman Test: Causal or Predictive

Although we have an efficient causal estimator, a common defect of causal estimators is large variance which implies poor predictive performance. By reducing the model's bias, we (perhaps expectedly) increase its variance, usually by a lot. This is the side effect of projecting our model's variables $X$ and $Y$ onto our instrumental variables, $Z$. There are two potential solutions to this problem: 1) create additional instrumental variables or otherwise increase their variance through more aggressive randomization[4] or 2) abandon causation in favor of correlation. Assuming that we have already created the maximum allowable variance in our instrumental variables (e.g., due to practical constraints such as budgets, user experience, etc.), we are obviously committed to causal estimation, so neither solution is practical. An ideal method would combine the advantages of both estimators. The Hausman test provides this.

The Hausman test is a frequentist model-selection test used to evaluate two models under a null hypothesis that both models are consistent but one is more efficient. For example, suppose $\hat{\beta}_{OLS}$ is efficient under the null, but only $\hat{\beta}_{IV}$ is consistent under the alternative that correlation is not causation: $E[\varepsilon'X] \neq 0$. The test then compares the two estimators in a special F test which recognizes $Cov(\hat{\beta}_{IV}, \hat{\beta}_{OLS}) = Var(\hat{\beta}_{OLS})$ and simplifies the denominator:

$$H = \left(\hat{\beta}_{IV} - \hat{\beta}_{OLS}\right)' \left(Var\left(\hat{\beta}_{IV}\right) - Var\left(\hat{\beta}_{OLS}\right)\right)^{-1} \left(\hat{\beta}_{IV} - \hat{\beta}_{OLS}\right).$$

This Hausman test statistic, $H$, has an asymptotic $\chi^2$ distribution with degrees of freedom equal to difference in rank between the two estimators' covariance matrices. The power of the Hausman test lies in its ability to let us abandon a noisy, high-variance causal estimate when the strength of the evidence of the causal estimator differing from the correlational estimator is weak. For example, if the coefficients of OLS and IV are statistically indistinguishable, we might be comfortable using the more precise OLS estimator because IV has bounded the magnitude of its bias.

## Hausman Penalization: Causally Consistent and Predictive

The astute student of machine learning might recognize that the Hausman test looks a lot like an offset $L^2$ penalty between the two estimators. This motivates our efficient causal estimator:

---

[4] We discuss a principled form of randomization in [Practical Exploration & Reporting: Bayesian Bootstrap](#). In the discussion, we present a tunable multiplier on the covariance matrix that simulates a reduction or expansion of the sample size in order to expand and contract, respectively, the variance of exploration via Thompson sampling.

causal estimation with Hausman penalization between the correlation estimator and the causal estimator. For example, per standard machine learning protocols, we would standardize our features and then $L^2$ penalize our efficient IV estimator toward ridge regression, rather than zero:

$$\hat{\beta}_{Hausman} \equiv argmin_\beta \; \hat{\varepsilon}(\beta)'g(Z)\bar{\Omega}^{-1}g(Z)'\hat{\varepsilon}(\beta) + \lambda_{Hausman}\|\beta - \hat{\beta}_{Ridge}\|^2$$

$$\text{where } \hat{\beta}_{Ridge} \equiv argmin_\beta \; \hat{\varepsilon}(\beta)'\hat{\varepsilon}(\beta) + \lambda_{Ridge}\|\beta\|^2$$

and $\bar{\Omega}^{-1} \approx Var(\varepsilon'g(Z))^{-1}$.

## Simple Hausman Penalization: Control Function via Ridge Regression

At this point, an astute econometrics student might recognize that a simple way to obtain a very similar and very simple-to-compute estimator for small-scale causal inference problems: use ridge regression to compute 2SLS via the control function approach, penalizing the control functions' coefficients. The control function approach is a simple way to both estimate 2SLS and execute the Hausman test using least squares.

Recall our 2SLS model equations:

$$Y = X\beta + \varepsilon \; ; \; X = Z\pi + \nu \; ; \; E[\varepsilon'X] = E[\varepsilon'\nu] \neq 0 \quad .$$

The control function approach estimates $Y = X\beta + \hat{\nu}\beta_{\hat{\nu}} + \varepsilon$ where $E[\varepsilon'X|\hat{\nu}] = 0$ and $\hat{\nu} = X - \hat{X} = X - Z\hat{\pi}_{OLS}$, resolving the endogeneity bias in $\hat{\beta}_{OLS}$. Testing $\hat{\beta}_{\hat{\nu}} = 0$ is an equivalent form of the Hausman test.

The key observation in this model is that $\hat{\beta}_{\hat{\nu}} = \hat{\beta}_{OLS} - \hat{\beta}_{2SLS}$. Hence, as we increase the penalization of $\hat{\beta}_{\hat{\nu}}$ from zero to infinity, $\hat{\beta}$ transitions from $\hat{\beta}_{2SLS}$ to $\hat{\beta}_{OLS}$. Finally, because $\hat{\nu}$ is a function of $Z$, we can select the magnitude of penalization by performing cross validation just as we normally would. For example, if our loss function strictly prefers unbiasedness (minimizing variance), we may find ourselves selecting the 2SLS (OLS) model. Mean-squared error loss will imply a convex combination of OLS and 2SLS.

**Control Function Approach to 2SLS (via Ridge Regression):**

1. Estimate OLS of $X$ on $Z$ to obtain $\hat{\nu} = X - \hat{X} = X - Z\hat{\pi}_{OLS}$.
2. Estimate OLS (Ridge on $\hat{\nu}$) of $Y$ on $X, \hat{\nu}$ to obtain $\hat{\beta}_{2SLS}$ and $\hat{\beta}_{\hat{\nu}}$.
3. Test $\hat{\beta}_{\hat{\nu}} = 0$ ($L^2$ penalize $\hat{\beta}_{\hat{\nu}}$) for the Hausman test (penalization).

The only difference between the control function approach and the Hausman penalization above is independent penalization of the Ridge coefficients from the Hausman penalty. This can be achieved in most statistical software by imposing different penalization weights on the Hausman penalty, depending on your preference for trading off bias and variance.

With a clear understanding of Hausman penalization in linear models, we now generalize the Hausman penalization.

## Hausman Causal Correction: Hausman Penalization in Practice

We motivate and generalize the concept of Hausman penalization as the "Hausman Causal Correction" (HCC). The HCC for linear models is simply the difference in model parameters, $\beta_{HCC} = \beta_{Hausman} - \beta_{Corr}$, or predictions, $\hat{Y}_{HCC} = \hat{Y}_{Hausman} - \hat{Y}_{Corr}$.

Industrial applications usually present an incumbent correlational machine learning model that works well enough for a business to be viable through many iterations of tuning and A/B testing, which is a slow and crude form of causal training. However, data scientists, engineers, and managers are collectively aware of the potential biases in the system---collectively known as endogeneity to the econometrician. Examples include winning or losing advertising auctions due to private information held by other bidders or organic feedback and path dependency from using today's model to generate the data to be used to train tomorrow's model. Frequently, these models are complex and high-dimensional black boxes maintained by large teams. It is unrealistic to suppose that an initial iteration of a causal model could (or even should) completely replace the incumbent model. However, even a simple causal model might identify first-order improvements in performance by taking the incumbent correlational model and making small, but rigorously estimated adjustments to the predictions, scores, rankings, or valuations. To this end, we describe the HCC estimator below, beginning with the first step of estimating the incumbent correlational machine learning model.

**Hausman Causal Correction (HCC) Estimator:**
1. Estimate the correlational model: $\hat{\beta}_{Corr}$ and $\hat{Y}_{Corr}$.
2. Estimate the causal model (e.g., GMM objective function) using Hausman penalization: $\lambda_{HCC}\|\beta - \hat{\beta}_{Corr}\|^2$ or $\lambda_{HCC}\|\hat{Y} - \hat{Y}_{Corr}\|^2$.
3. Cross-validate $\lambda_{HCC}$ by minimizing the GMM objective function or some other loss out of sample.

Upon estimating the correlational model, the second step is to impose the Hausman penalization, either implicitly or explicitly. In linear models with default penalization toward zero, this can be done implicitly by first subtracting out the prediction $\tilde{\varepsilon} = Y - X\hat{\beta}_{Corr}$ (e.g., estimating the causal model on the correlational model's residual) or introducing a linear index model prediction offset $h(X\beta - X\hat{\beta}_{Corr})$ for a nonlinear link function, $h(\cdot)$. In more complex models with default penalization toward zero, it requires offsetting/recentering the model parameters by the correlational model's estimates (e.g., optimize with respect to $\delta_{Causal} = \beta - \beta_{Corr}$) or explicitly imposing the Hausman penalization. For more general models, we can perform a "regression on the residual" (RoR) with regularization on the prediction: $\tilde{\varepsilon} = Y - \hat{Y}_{Corr}$ and a penalty of $\lambda_{HCC}\|\hat{Y} - \hat{Y}_{Corr}\|^2$. Such a hybrid correlation + causation model will then have marginal effects of $\frac{\partial \hat{Y}_{Hausman}}{\partial X} = \frac{\partial \hat{Y}_{Corr}}{\partial X} + \frac{\partial \hat{Y}_{HCC}}{\partial X}$ or, more specifically for our linear IV model, $\frac{\partial \hat{Y}_{Hausman}}{\partial X} = \frac{\partial \hat{Y}_{Corr}}{\partial X} + \hat{\beta}_{HCC}$.

In the third and final step, cross validation determines the strength of the causal model's evidence for deviations from the correlational model. Under strong preferences for unbiasedness, $\hat{\beta}_{Corr}$ may simply serve as a predictive numerical stabilization of the causal model (e.g., correlational ridge prior, rather than priors of zero). Under strong preferences for low variance, $\hat{\beta}_{Causal}$ is only used to nudge the correlational model away from the most egregious errors and perhaps encourage slow drift toward stricter causality over time.

A simple implementation of estimating the HCC using our linear IV causal model is what we informally call the "regression on the residual" (RoR) by first subtracting out the correlational model's prediction before estimating a regularized (to zero) causal model.

**HCC via "Regression on the Residual" (RoR) for Linear Models:**
1. Estimate the correlational linear model: $\hat{\beta}_{Corr}$. Cross-validate any hyperparameter using mean-squared error. Compute the residual, $\hat{\varepsilon}_{Corr} = Y - X\hat{\beta}_{Corr}$.
2. Estimate a zero-penalized, $\lambda_{HCC}\|\hat{\beta}_{HCC}\|^2$, IV model of the residual $\hat{\varepsilon}_{Corr}$ on $X$ instrumented by $Z$.
3. Cross-validate $\lambda_{HCC}$ using the GMM objective function, $\lambda_{HCC} = min_\lambda \hat{\varepsilon}_{IV}(\hat{\beta}_{HCC})'ZZ'\hat{\varepsilon}_{IV}(\hat{\beta}_{HCC})$, over the cross-validation sample.

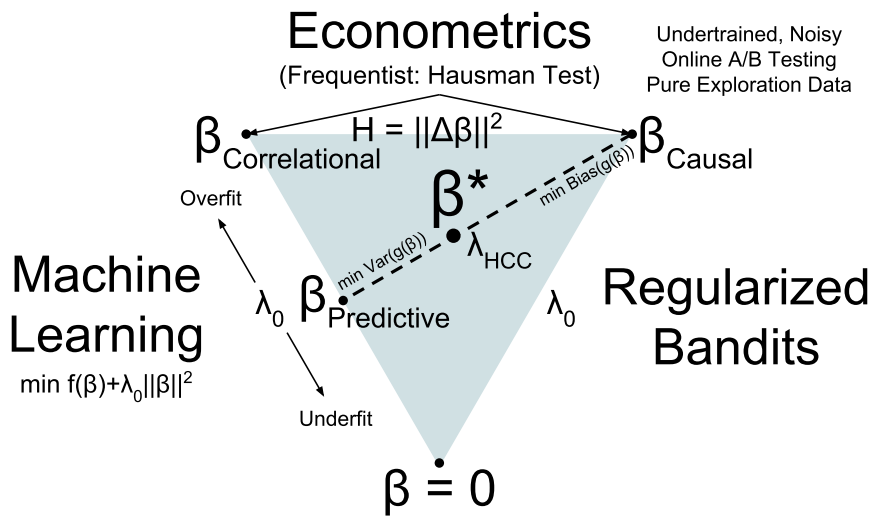

Figure 9 - Hausman Causal Correction and Conventional Machine Learning

In Figure 9, we provide intuition for the HCC within the context of econometrics, standard correlational machine learning, and regularized bandits. In particular, the HCC is simply a continuous convexification of the space between the discrete choices of correlational and causal econometric models spanned by the Hausman test and the space between the zero-prior model selection methods commonly used in predictive and bandit machine learning applications. The primary advantage of the Hausman Causal Correction is its ability to provide a finite-sample causal refinement to any correlational model. This is especially useful when our loss function depends on more than consistency or bias of our parameter estimates.

## HCC Simulation Study: Efficient and Consistent Bandits

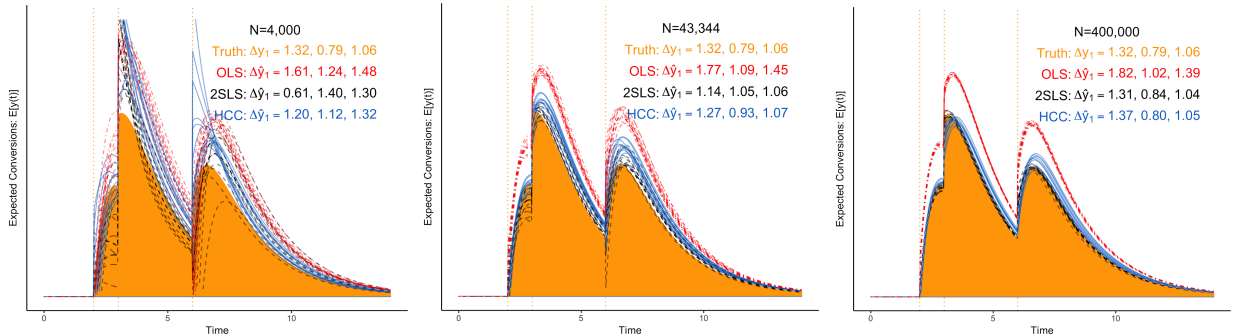

Figure 10 - Consistency of Hausman Penalization & Causal Correction

Lewis & Wong (2018) provide a more thorough treatment of the Hausman Causal Correction in the setting of econometric bandits. In Figure 10, we give a simple illustration of the distribution of estimates for OLS, 2SLS, and HCC for sample sizes of N=4,000; N=43,344; and N=400,000. As the sample size grows, the HCC estimates drift from approximating OLS to 2SLS.

While the regularized control function approach is a fine causal model for low-dimensional problems by balancing the tradeoff between correlational and causal models, its computation is non-scalable because $\hat{v}$ is dense (not sparse). As such, we need a scalable IV model that can accommodate a large number of endogenous variables for the purpose of obtaining causal estimates of heterogeneous treatment effects.

## Scalable Estimation of Sparse Instrumental Variables Models

IV is a special case of the generalized method of moments (GMM) estimation framework. GMM takes a moment condition, $E[g(\beta)|Y,X,Z] = 0$, and minimizes its weighted quadratic form

$$\hat{\beta}_{GMM} \equiv argmin_{\beta} \ \hat{g}(\beta)'\Omega^{-1}\hat{g}(\beta) + \lambda_{GMM}\|\beta\|^2$$

where the weighting matrix $\Omega^{-1} = Var(g(\beta))^{-1}$ minimizes the variance of the estimator. Using a plug-in estimator of $\tilde{\Omega}^{-1} \approx Var(\hat{g}(\beta))^{-1}$ produces a feasible estimator known as optimal GMM. In practice, a two-step estimation strategy using $\Omega = I$, the identity matrix, in a first round of minimization facilitates a second round of minimization using the plug-in estimator of the weighting matrix. Here we deviate from standard GMM practice by imposing a quadratic penalty on the model parameters. Linear IV's GMM moment condition is the conditional independence assumption: $E[g(\beta)] = E[\varepsilon'Z] = 0$ where $\varepsilon = Y - X\beta$. $\Omega = Z'Z$ yields 2-stage least squares (2SLS) and $\Omega = Z'\Sigma Z$ where $\Sigma = E[\varepsilon\varepsilon']$ yields the generalized least squares (GLS) version of 2SLS, optimal or efficient IV. In the case where $Z = X$, these models reduce to ordinary least squares (OLS) and GLS. For optimal IV, we explicitly write out the GMM objective function:

$$\hat{\beta}_{IV} \equiv argmin_\beta \ \hat{\varepsilon}(\beta)'Z\bar{\Omega}^{-1}Z'\hat{\varepsilon}(\beta) + \lambda_{IV}\|\beta\|^2.$$

We seek a computationally scalable version of this objective function.

First, why is optimal IV not scalable? We provide two primary reasons: the objective function and its first order condition include a large matrix inversion and elements that are quadratic in the number of features. We focus our discussion on the first order condition:

$$X'Z\bar{\Omega}^{-1}Z'(Y - X\beta) = \lambda I \ .$$

The matrix inversion problem can be readily "solved" by $\bar{\Omega}^{-1} \approx Var(Z\Sigma Z)^{-1} = V_Z^{-1}$ where $V_Z^{-1}$ is a sparse approximation, $O(columns(Z))$, trivially invertible diagonal matrix or directly tuned via cross validation to minimize the out-of-sample GMM objective function's quadratic form and, hence, variance. In the diagonal case, we can simply replace $Z$ with $\tilde{Z} = ZV_Z^{-\frac{1}{2}}$. We have eliminated the matrix inversion by explicitly trading off second-order statistical efficiency and computational scalability. In practice, using a diagonalized covariance matrix in sparse optimization problems provides a suitable first-order approximation because most off-diagonal covariances are either identically or approximately zero. The $L^2$ ridge regularization $\lambda I$ implies a similar overemphasis of a diagonal covariance. In exactly identified IV problems, the diagonalization of the covariance matrix has no effect on the theoretical estimates, but provides numerical stability to the computation of the estimator.

Even after eliminating the matrix inverse, we still have computational complexity that is $O(columns(Z)^2)$ from $\tilde{Z}\tilde{Z}'$. Jebara & Rostykus (2018) propose solutions to this problem by trading off second-order statistical efficiency for a reduction in computational complexity. They do this by using importance sampling proportional to the magnitude of the sparse instrumental variables. This clever trick prioritizes the computation of the largest gradient steps while still maintaining the overall expectation. We defer more discussion of their method until a future draft of this paper and theirs. In practice, the hierarchy of the heterogeneous ad stock regressors and their corresponding instrumental variables makes $\tilde{Z}\tilde{Z}'$ and $X'\tilde{Z}$ sparse, so we can simply hard code and directly exploit that sparsity.

## Solution: Scalable IV with Hausman Penalization via HCC

The production-ready causal machine learning valuation algorithm satisfies our requirements:
- **Causal**: Linear IV.
- **Predictive**: Hausman Penalization via HCC.
- **Scalable**: Estimation via SGD IV (>5,000 features) or control function approach using PCG (<=5,000 features).

- **Efficient**: Large Scale Sparse Designer IVs + Feasible Optimal 2-Step GMM.

**Scalable IV with Hausman Penalization via HCC:**

1. Estimate correlational model (e.g., ridge regression): $\hat{\beta}_{Corr}$. Compute $\hat{\varepsilon}_{Corr} = Y - \hat{Y}_{Corr}$.
2. Estimate penalized linear IV on the residual using either SGD IV or the control function approach via PCG: $\hat{\beta}_{HCC}$.
3. Cross-validate the HCC penalization, $\lambda_{HCC}$, using the linear IV GMM objective function using a holdout sample.
4. Return $\hat{\beta}_{Hausman} = \hat{\beta}_{Corr} + \hat{\beta}_{HCC}$ as the final model estimates for computing incrementality.

This solution provides a simple, scalable approach to refine an incumbent model in accordance with the strength of the evidence that causal decision-making materially changes the correlational model's causal predictions. As evidence builds that correlation is not causation, the model automatically updates.

# Practical Exploration & Reporting: Bayesian Bootstrap

Lewis & Wong (2018) broadly discuss econometric bandits; we briefly describe their generalization and practical implementation of Thompson sampling using the Bayesian bootstrap.

## Bayesian Bootstrap for Confidence Intervals & Exploration

For econometric estimators satisfying the standard requirements for asymptotic normality, we can utilize the nonparametric bootstrap to simultaneously draw from and approximate the distribution of the estimator. We recommend using a small number of bootstrap draws (e.g., 10 or 20) using the Bayesian bootstrap to simplify computation and facilitate Thompsons Sampling from even high-dimensional posteriors. We can sample from the posterior of our estimator by randomizing for each bid which of the Bayesian bootstrap draws we use to compute the expected value of an impression.

If we want to continue to learn, we need our implementation of Thompson sampling to lead us to sometimes win or lose the auction. If there is a single clearing price that we almost always exceed, the Hausman Causal Correction may not be consistent. Enforced 2-point (e.g., zero and non-zero) bidding can ensure an identified first-stage by both winning and losing. In short, our exploration should reflect our uncertainty of both the value of the impression and our confidence about the value being > 0.

In general, if we believe that our current sampling method for the Bayesian bootstrap understates the true variance of the estimator, we can adapt the Bayesian bootstrap's sampling distribution to have higher variance. This simulates larger variance analogous to decreasing the sample size.

## Bonus: Bayesian Bootstrap for Cross Validation

The bootstrap can be used for cross validation and model selection. As such, since we are already running Bayesian bootstraps for standard errors, we can also use those same samples to perform cross validation. This is done by simply evaluating the cross validation loss function using all of the Bayesian bootstrap estimates against the entire sample and choosing the hyperparameters that minimize the loss function. We defer a more complete discussion of these ideas to a future draft.

# Conclusion

Incrementality answers the advertiser's central question: "What is the causal effect of my ads?" To the end of answering that question, we have proposed a method that encompasses the ad impression valuation/bidding/buying and attribution problem using rigorous causal methods powered by randomization mechanisms, similar to Johnson, Lewis, & Nubbemeyer's (2017) "Predicted Ghost Ads." The method can be simply summarized: we use a regularized instrumental variable (IV) model of heterogeneous treatment effects in a continuous-time panel. We utilize the Bayesian bootstrap as a computationally simple approach to compute standard errors, perform model selection, and approximate Thompson sampling. While most of the components of the model may not be novel, the viability of their combination represents a step forward for incrementality bidding.